\documentclass[10pt,twocolumn,letterpaper]{article}

\usepackage[pagenumbers]{cvpr} 
\definecolor{cvprblue}{rgb}{0.21,0.49,0.74}

\usepackage{multirow}
\usepackage[pagebackref,breaklinks,colorlinks,allcolors=cvprblue]
{hyperref}

\def\paperID{} 
\def\confName{CVPR}
\def\confYear{2026}

\title{CARE: A Molecular-Guided Foundation Model with Adaptive Region Modeling for Whole Slide Image Analysis}
\author{
Di Zhang$^{1}$ \quad Zhangpeng Gong$^{1}$\thanks{Zhangpeng Gong and Xiaobo Pang contributed equally to this work.} \quad Xiaobo Pang$^{1}$\footnotemark[1] \qquad Jiashuai Liu$^{1}$ \quad Junbo Lu$^{1}$ \quad Hao Cui$^{1}$ \\
Jiusong Ge$^{1}$ \quad Zhi Zeng$^{1}$ \quad Kai Yi$^{2}$ \quad Yinghua Li$^{3}$ \quad Si Liu$^{3}$ \qquad Tingsong Yu$^{3}$ \quad Haoran Wang$^{4}$ \\
Mireia Crispin-Ortuzar$^{2}$ \quad Weimiao Yu$^{5}$ \quad Chen Li$^{1}$\thanks{Co-corresponding authors. $<$cli@xjtu.edu.cn, zg323@cam.ac.uk$>$} \quad
Zeyu Gao$^{2}$\footnotemark[2] \\
{
$^{1}$Xi'an Jiaotong University \quad
$^{2}$University of Cambridge \quad
$^{3}$KingMed \quad
$^{4}$BGI Research \quad
$^{5}$A$^\star$STAR
}
}

\begin{document}
\maketitle
\begin{abstract}
Foundation models have achieved success in computational pathology, demonstrating generalization across histopathology tasks. However, existing models overlook the heterogeneous and non-uniform organization of regions of interest (ROIs) because they rely on natural image backbones not tailored for tissue morphology. Consequently, they fail to capture the coherent tissue architecture beyond patches, limiting interpretability and clinical relevance. To address these challenges, we present Cross-modal Adaptive Region Encoder (CARE), a foundation model for pathology that partitions WSIs into several morphologically relevant regions. Specifically, CARE employs a two-stage pretraining strategy: (1) a self-supervised unimodal pretraining stage that learns morphological representations from 34,277 whole-slide images (WSIs) without segmentation annotations, and (2) a cross-modal alignment stage that leverages RNA and protein profiles to refine the construction and representation of adaptive regions. This molecular guidance enables CARE to identify biologically relevant patterns and generate irregular yet coherent tissue regions, selecting the representative area as ROI. CARE supports a broad range of pathology-related tasks, using either the ROI feature or the slide-level feature obtained by aggregating adaptive regions. Based on only one-tenth of the pretraining data typically used by mainstream foundation models, CARE achieves superior average performance across 33 downstream benchmarks, including morphological classification, molecular prediction, and survival analysis, and outperforms other foundation model baselines overall. 
\end{abstract}    
\section{Introduction}
\label{sec:intro}
Benefiting from advances in deep learning, computational pathology (CPath) is increasingly embedded in clinical workflows \cite{song2023artificial}. Histopathology remains the diagnostic gold standard for major diseases such as cancer, commonly read as digitized whole-slide images (WSIs). Leveraging WSIs, CPath algorithms perform quantitative analyses for tumor detection \cite{shi2024vila}, disease characterization \cite{lu2024multimodal}, and prognostic assessment \cite{chen2024predicting}. Building on this digital substrate, pathology foundation models \cite{wang2022transformer} trained with self-supervised learning on large-scale collections of pathology patches have recently emerged. Compared with models pretrained on natural images, they learn domain-specific representations that transfer effectively to diverse downstream tasks and yield stronger performance, particularly when only limited labeled data are available.

\begin{figure}[t]
  \centering
   \includegraphics[width=\linewidth]{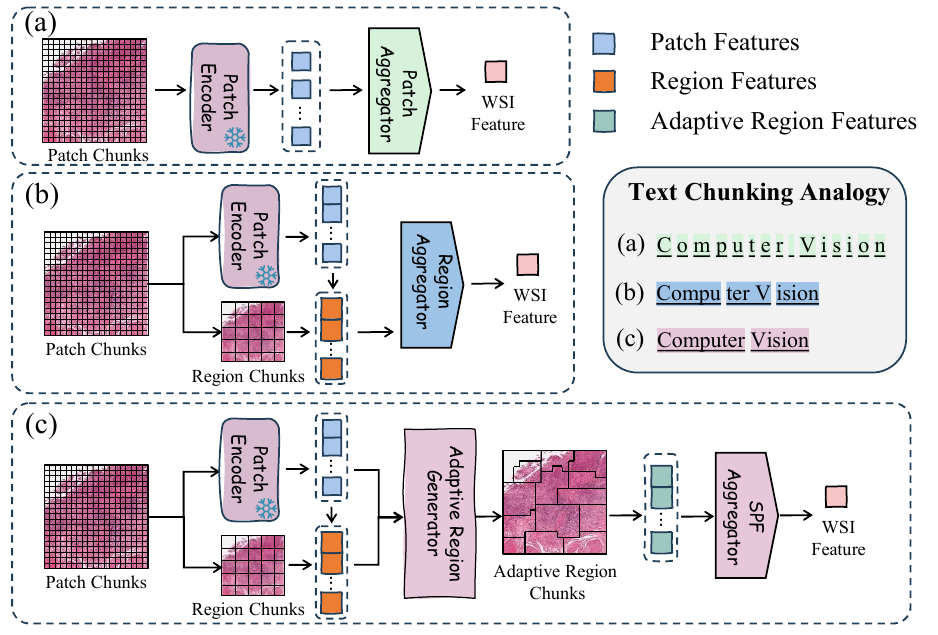}

   \caption{CARE \vs conventional CPath chunking on WSIs. Panels (a) and (b) depict patch chunks and regular region chunks, respectively. They both impose artificial grids that split tissue, making them inefficient and semantically weak. Panel (c) shows our adaptive region chunks (CARE), which behave like word-level tokens, respect tissue boundaries, and capture meaningful texture, morphology, and cell layout.}
   \label{motivation}
\end{figure}

Pathology image analysis has traditionally centered on processing WSIs. 
Advances in patch-level foundation models have, in turn, enabled slide-level models that learn holistic embeddings of entire WSIs. 
This shift broadens downstream applicability and improves inference efficiency for gigapixel-scale pathology slides. With the integration of self-supervised pretraining \cite{simeoni2025dinov3} and multimodal contrastive learning approaches \cite{radford2021learning,yu2022coca}, slide-level foundation models \cite{chief,xu2024whole} can better exploit the information in pathology slides. As a result, they serve as powerful tools for tasks such as visual analysis, pathology report generation, modality fusion and retrieval, and zero-shot classification.

Most current slide-level foundation models in computational pathology inherit natural-image backbones and treat WSIs as large collections of fixed-size patches. In routine diagnostics, however, pathologists first localize diagnostically significant ROI. By contrast, typical foundation models lack an explicit ROI concept. They either apply global self-attention \cite{titan,prism} or linear attention over all patches \cite{abmil} or impose fixed, rule-based partitions \cite{hipt,rrt}. As summarized in \cref{motivation}, the prevailing chunking schemes are patch chunks (\cref{motivation}(a)) or regular region chunks (\cref{motivation}(b)). This mirrors tokenization in natural language processing: patch chunks are like character-level tokens---fine-grained yet myopic and regular regions are like fixed-length chunking---broader yet prone to semantic misalignment. Most models aggregate information directly from \emph{patches} to \emph{WSI}, creating long-range interactions across thousands of patches and making the aggregator hard to learn. A hierarchical from \emph{patch} to \emph{region} to \emph{WSI} scheme would better align with tissue structure, help avoid long-range interactions, and improve data efficiency. However, existing region chunking remains crude, which is why \emph{patch–region–WSI} backbone has not become mainstream.

To address these limitations, we introduce CARE (Cross-modal Adaptive Region Encoder), a pathology foundation model that replaces rigid patch or region chunks with adaptive region chunks, analogous to word-level tokenization in NLP. Each adaptive region partition is a semantically coherent unit that captures an entire tissue structure (\cref{motivation}(c)), supplying the model with structure-aware context that conventional slide-level pipelines typically miss. 
CARE includes an adaptive region generator (ARG) that automatically discovers and partitions morphologically relevant regions in WSIs, thereby improving semantic expressiveness and interpretability. To pretrain CARE, we adopt a two-stage strategy that uses molecular data to guide and refine regionalization. The first stage applies Image BERT Pre-Training with Online Tokenizer (iBOT) \cite{ibot} on WSIs in a purely self-supervised manner, producing robust initial weights that enable data-efficient molecular alignment in the subsequent stage. The second stage performs cross-modal alignment with matched molecular profiles so that region embeddings reflect underlying biology: we first align slides with RNA expression profiles for broad, high-coverage supervision, and then align them with protein expression profiles to incorporate higher-specificity signals. This curriculum sharpens region boundaries and yields representations that are both tissue-aware. Importantly, we employ a lightweight aggregator that performs weighted pooling across adaptive regions, implicitly aligning molecular expression with biologically relevant tissue regions. We evaluate the pretrained model on 33 downstream tasks. CARE attains superior performance while using only about one-tenth or less of the pretraining data used by mainstream pathology foundation models, underscoring the effectiveness of adaptive region modeling in computational pathology. Our contributions are as follows:

\begin{itemize}
	
	\item We present CARE, a pathology foundation model. After self-supervised pretraining, we align slide features to RNA/protein profiles to refine adaptive region partitioning and representations, reducing the labeled data requirement for foundation model training to about one-tenth of that typically needed.
	\item We propose an adaptive region generator that partitions WSIs into irregular yet morphologically relevant regions, aligns region construction with pathologists’ ROI-centric workflow, and enables both ROI-level analysis and slide-level prediction through region feature aggregation.
	\item CARE delivers strong results on 33 downstream tasks. Across linear probing and fine-tuning, CARE’s adaptive region design and pretraining pipeline consistently outperform prior approaches.
\end{itemize}


\section{Related Works}
\label{sec:formatting}

\noindent\textbf{Computational Pathology Analysis.} CPath aims to infer diagnostic, prognostic, and molecular signals from WSIs, and has shifted from handcrafted textures to weakly supervised end-to-end learning with multiple instance learning (MIL) \cite{cersovsky2023towards, li2025mico} as the dominant paradigm. ABMIL \cite{abmil} aggregates patch features via attention to produce slide-level predictions. {HIPT} \cite{hipt} introduces a region-level stage that consolidates adjacent and semantically homogeneous patches with a hierarchical Transformer before global fusion. Variants such as DSMIL \cite{li2021dual} and DTFD-MIL \cite{zhang2022dtfd} further refine patch interactions and leverage distillation. Building on these advances, the scope of CPath now extends beyond diagnosis. It covers time-to-event modeling \cite{yao2020whole} and prediction of treatment and immune response \cite{li2023outcome}. It also spans spatial transcriptomics, from cross-modal integration \cite{he2020integrating} and high-resolution histology-based inference \cite{shi2024high} to scalable WSI generation via flow matching \cite{huang2025scalable}.

\noindent\textbf{Patch-level Foundation Model in Pathology.} Modern pathology representation learning proceeds from \emph{patch} to \emph{WSI}. Patch-level encoders capture cellular and local morphology with self-supervised pretraining \cite{wang2023retccl,yan2022deep, li20252}. In vision–language modeling, {PLIP} \cite{huang2023visual}, {CONCH} \cite{lu2024visual}, and {BiomedCLIP} \cite{zhang2023biomedclip} align tiles with text yet remain patch-centric. By scaling unimodal self-supervision, UNI \cite{uni} yields a versatile patch encoder with strong transfer. {MUSK} \cite{musk} similarly provides a tile-focused vision–language encoder that attains competitive slide performance when paired with MIL aggregators. Collectively, these approaches are patch-level foundation models and do not support end-to-end diagnostic inference from WSIs. This gap motivates slide-level foundation models in pathology.


\noindent\textbf{Slide-level Foundation Model in Pathology.} In response, researchers pretrain slide encoders at the \emph{WSI} level that operate directly on whole slides. {CHIEF} \cite{chief} injects anatomy-aware attention under weak supervision. {Prov\mbox{-}GigaPath} \cite{xu2024whole} pairs a ViT tile encoder with a \emph{LongNet}-based slide encoder and uses masked-tile objectives to handle ultra-long sequences. {TITAN} \cite{titan} constructs a coordinate-aware feature grid and aggregates features with slide-level Transformers. {Feather} \cite{feather} shows that a lightweight ABMIL encoder can be pretrained at the slide level and transfers well. {PRISM} \cite{prism} learns encoders supervised by clinical reports and dialogues to align slide embeddings with diagnostic semantics. In a complementary direction, {TANGLE} \cite{tangle} aligns WSIs with transcriptomic profiles during pretraining to inject molecular guidance. The emerging direction couples strong patch encoders with pretrainable slide-level aggregators, enabling fully automated slide-level reasoning.


\section{Methodology}
\label{sec:method}

\begin{figure*}
  \centering
  \includegraphics[width=\linewidth]{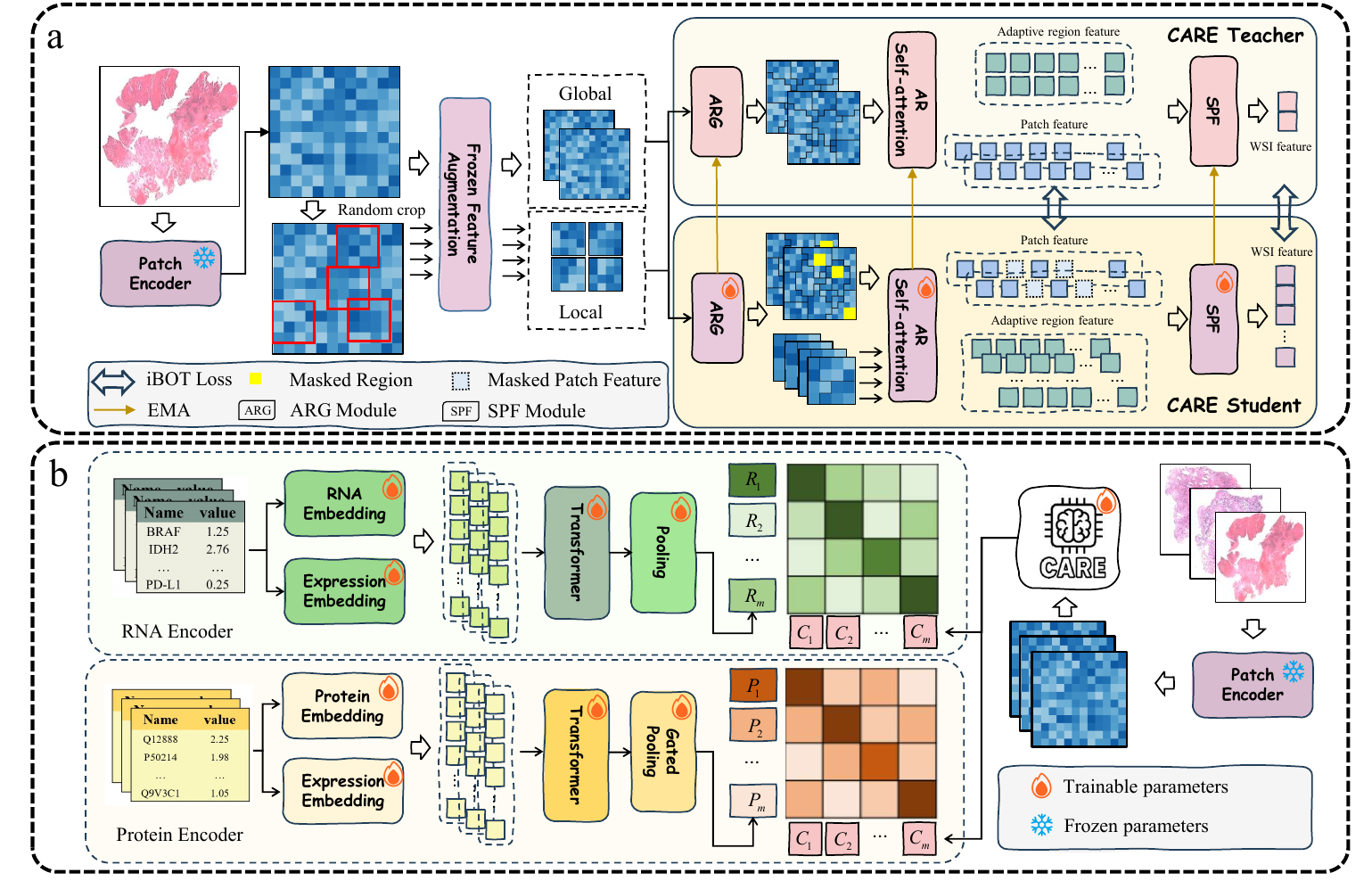}
  \caption{CARE architecture and training pipeline. (a) CARE framework with iBOT-style self-supervised pretraining using a teacher–student setup. CARE comprises three modules: the Adaptive Region Generator (ARG), adaptive region self-attention (ARSA), and Semantic and Prior Fusion (SPF). ARG partitions WSIs into morphologically coherent regions. ARSA operates within each adaptive region to derive region level features. SPF aggregates these features into a WSI embedding. (b) Cross-modal pretraining pipeline. RNA and protein encoders produce molecular embeddings, which are aligned to CARE’s WSI embedding via an InfoNCE loss.}
  \label{framework}
  \vspace{-0.2\baselineskip}
\end{figure*}
\subsection{Preliminaries}

Due to the gigapixel scale of WSIs, end-to-end training at native resolution is impractical. Thus, most methods partition slides into fixed-size patches, extract patch features, and aggregate them into a slide-level embedding.

Firstly, we perform tissue segmentation and then tile patches for WSI. A variety of patch-level feature extractors can model the fine-grained morphology of histopathology images. We adopt CONCH v1.5 \cite{lu2024visual} as our patch feature extractor, which is a patch-level pathology foundation model widely used in practice.  Given a WSI $I\in\mathbb{R}^{H\times W\times 3}$ scanned at $20\times$ magnification, we extract $N$ square patches by sliding a $p\times p$ window. Let $\mathbb{P} = \{P_i =(u_i,v_i,p,f_i) \}_{i = 1}^N$ denote the set of patches, where $(u_i,v_i)$ denotes the top-left coordinate  and $f_i\in \mathbb{R}^d$ denotes the feature representation produced by patch encoder. 

Secondly, slide-level foundation models or MIL models operate on a set of patch embeddings and learn a slide encoder to aggregate instance features into a slide-level representation for tasks such as diagnosis, prognosis, and retrieval. In this study, we present CARE, a slide-level foundation model tailored for histopathology that restructures the input from raw patches into adaptive regions.

\subsection{Model Design}


\subsubsection{Subregion Partition and Representation}

Given the patch set $\mathbb{P}$, we define $M$ non-overlapping square subregions on the patch grid, $\mathbb{S} = \{S_i\}^M_{i = 1}$, where $S_i=(x_i,y_i,k,g_i^{\mathrm{CLS}},g_i^{Q})$ and $(x_i,y_i)$ is the top-left patch and $k\in\mathbb{N}$ is the window side length (in patches). $(g^\mathrm{CLS}_i,g^{Q}_i)$ denote the CLS-aggregated and query-aggregated features of $S_i$, respectively. Each $S_i$ covers a $k\times k$ block anchored at $(x_i,y_i)$, \ie, patches $(u,v)$ with $x_i\le u<x_i+k$ and $y_i\le v<y_i+k$. $g_i^{\mathrm{CLS}}$ and $g_i^{Q}$ are obtained by aggregating patch features within the window, shown in \cref{module}(a).

\noindent\textbf{CLS-aggregated subregion feature.} We apply multi-head regional self-attention within each subregion. Let $F_i$ be the sequence of patch embeddings in $S_i$. We prepend a learnable $[\mathrm{CLS}]$ and feed $([\mathrm{CLS}],F_i)$ into the regional self-attention layer. The $[\mathrm{CLS}]$ output is the subregion feature $g_i^{\mathrm{CLS}}$, and the remaining outputs form the subregion-aware patch features $F^R_i = \{f^R_{i,1},\cdots,f^R_{i,|S_i|}\}$.

\noindent\textbf{Query-aggregated subregion feature.} We introduce a learnable query $q\in \mathbb{R}^{1\times d}$ and apply multi-head regional cross-attention, \ie, standard cross-attention restricted to the patches within subregion $i$. Concretely, $q$ attends only to the patch embeddings of subregion $i$ to produce the query-aggregated subregion feature $g^Q_i$. This differs from global cross-attention in that the attention domain is confined to a single subregion rather than the entire slide.


\noindent\textbf{Soft inclusion.} We quantify subregion--patch relations with an anchor-distance soft inclusion. For patch $j$ with anchor $t_j=(u_j,v_j)$ and subregion $i$ with top-left anchor $s_i=(x_i,y_i)$, let $d_{ij}=\lVert t_j-s_i\rVert_2$. The soft inclusion is
\begin{equation}
  \mu_{ji} = 1-\dfrac{d_{ij}}{\max_{0 \leq j^\prime \leq N}d_{ij^\prime}}.
  \label{dis}
\end{equation}

The soft inclusion matrix is defined as $C =[\mu_{ji}]_{N*M}$.

\subsubsection{Adaptive Region Partition and Feature}

To break away from rigid, fixed-grid subregions and delineate irregular, characteristic-adaptive regions, we introduce an adaptive region generator that explicitly models spatial and feature relations between patches and subregions, as shown in \cref{module}(a).

\noindent\textbf{Adaptive region generator.} We reassign each patch to an adaptive region based on the soft inclusion matrix $C$. For patch $j$, we first select the indices of its top-$K$ candidate subregions ($K=3$ in this paper) ${\mathcal{I}_j = \operatorname{Top3}_{i\in\{1,\cdot,M\}}\big(C_{ji}\big)}$, where $i$ is the index corresponding to the subregion. Let the cosine similarity be $\cos(a,b)=\tfrac{a^\top b}{|a||b|}$. For each $i\in \mathcal{I}_j$, we compute the pairwise cosine similarities between the two patch representations and the two subregion descriptors, and stack these values into a vector $c_{ji} = [\cos(f_j,g^{\mathrm{CLS}}_i), \cos(f_j,g^{\mathrm{Q}}_i), \cos(f^R_j,g^{\mathrm{CLS}}_i), \cos(f^R_j,g^{\mathrm{Q}}_i)]$.

We then apply a per-patch softmax over the candidate set $\mathcal{I}_j$ for each score and average the resulting probabilities:
\begin{equation}
	\label{relat_ana2}
	\tilde{s}^{(m)}_{ji}=\frac{\exp^{s^{(m)}_{ji}}}{\sum_{i'\in\mathcal I_j} \exp^{s^{(m)}_{ji^\prime}}}, \quad
	\rho_{ji}=\dfrac{1}{4}\sum_{m = 1}^{4}\tilde{s}^{(m)}_{ji}.
\end{equation}

Finally, by combining semantic affinity with spatial proximity, we obtain the regional selection score $w_{j i}=\rho_{ji}\cdot C_{ji}$ for $i\in\mathcal{I}_j$. We then assign patch $j$ to the best candidate $A^{\star}(j)=\arg\max_{i\in\mathcal{I}_j} w_{ji}$. The resulting adaptive regions are defined as ${A}_i = \{j\vert A^{\star}(j) = i\}.$

\noindent\textbf{Adaptive region feature.} Let $m_i=\lvert A_i\rvert$ denote the cardinality of $A_i$. We then form the feature sequence $F_i^{\mathrm{AR}} = \{ f_j \vert j\in A_i\}$, where $f_j\in \mathbb{R}^{ d}$. We prepend a learnable classification token $[\mathrm{CLS}]_\mathrm{AR}\in \mathbb{R}^{d}$ and feed the concatenated sequence into adaptive region self-attention (ARSA), \ie,
\begin{equation}
	\label{arfea}
	\big[\,g_i^{\mathrm{AR}};\; F_i^{\mathrm{AR,cep}}\,\big] = \mathrm{ARSA}\left( [\mathrm{CLS}]_{\mathrm{AR}} \Vert F_i^{\mathrm{AR}} \right),
\end{equation}
where $g_i^{\mathrm{AR}} \in \mathbb{R}^d$ is the adaptive region representation given by the output at the $[\mathrm{CLS}]$ position, and $F_i^{\mathrm{AR,cep}} = \{ f^{\mathrm{cep}}_j \vert j\in A_i\}$ are the context-enriched patch features produced by self-attention within the adaptive region.

\noindent\textbf{Region structuring loss.} To prevent all patches from consistently selecting the nearest region during training and avoid collapse into a single region, we propose a new loss function, termed Region Structuring Loss (RSL).


For each patch $j$ and candidate subregion $i$ with selection score $w_{ji}$, we define the candidate rank $r_j(i)\in\{0,1,\ldots,K-1\}$. Here, $r_j(i)$ is the 0-based rank within the top-$K$, sorted by $w_{ji}$ in descending order (0 = largest). We then average over all valid patches to obtain the global mean expected rank: 
\begin{equation}
	\label{RERloss}
	\begin{aligned}
	\overline{E} &= \mathbb{E}_{j\sim\text{Unif}\{1,2,\cdots,N\}}\mathbb{E}_{i\sim{\pi_j}}[r_j(i)] \\
	&=\dfrac{1}{N} \sum_{j\in \{1,2,\ldots,N\}}\sum_{i\in\mathcal{I}_{j}}w_{ji}r_j(i),
	\end{aligned}
\end{equation}
where $\text{Unif}(\cdot)$ denotes the uniform distribution and $\pi_j$ denotes the regional selection score distribution for $j$-th patch.

We then pull $\overline{E}$ toward a target value ${E^\star}$ via $\mathcal{L}_\mathrm{RSL} = (\overline{E}-E^\star)^2$. The total training objective is 

\begin{equation}
	\label{sumloss}
	\mathcal{L}_\text{total} = \mathcal{L}_\text{main}+ \lambda_\mathrm{RSL}\mathcal{L}_\mathrm{RSL},
\end{equation}
where $\mathcal{L}_\text{main}$ is the primary task loss used during pre-training or fine-tuning and $\lambda_\mathrm{RSL}$ controls the strength of RSL.

\subsubsection{Slide-level Feature Representation}
\begin{figure}[t]
  \centering
   \includegraphics[width=\linewidth]{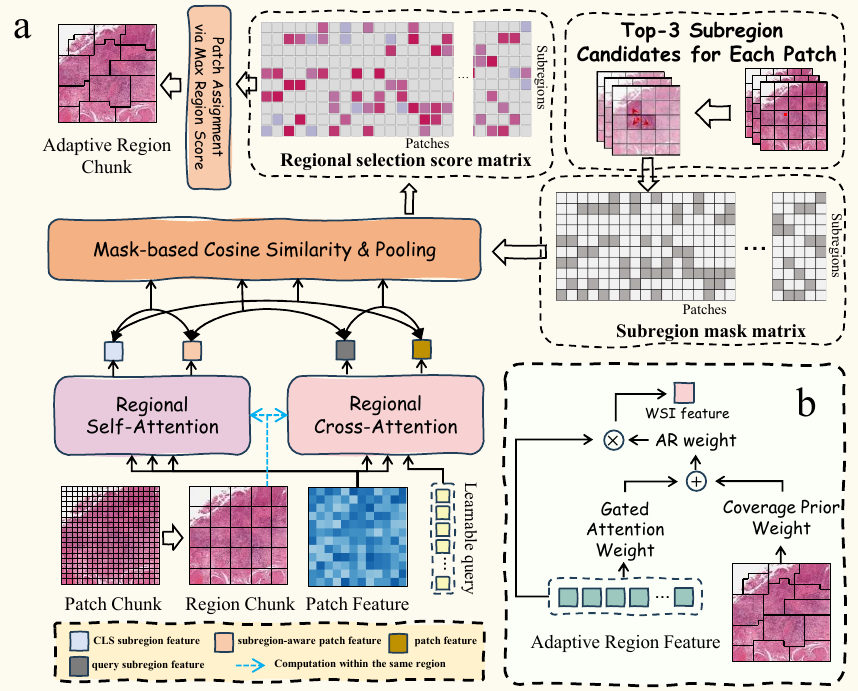}

   \caption{(a) Adaptive Region Generator. Based on soft inclusion, each patch retains only its top-3 candidate subregions and masks out the rest. Cosine similarity is then computed to the unmasked candidates, and the patch is assigned to the highest-scoring subregion, yielding an adaptive repartition of patches. (b) Semantic and Prior Fusion. A lightweight module that aggregates adaptive region features into a slide-level embedding.}
   \label{module}
   \vspace{-0.2\baselineskip}
\end{figure}

To pool the set of adaptive region features $\{g_i^{\mathrm{AR}}\}_{i=1}^M$ into a slide-level embedding, we propose Semantic and Prior Fusion (SPF) module (\cref{module}b), which fuses a coverage prior with semantic attention. The coverage prior $\alpha_i$ is the patch proportion of adaptive region $i$, and the semantic weight $\beta_i$ is computed via gated attention \cite{abmil}, \ie,
\begin{equation}
	\label{beta}
s_i=w^\intercal\big(\tanh(Vg_i^{\mathrm{AR}})\odot\sigma(Ug_i^{\mathrm{AR}})\big), \beta_i=\frac{e^{s_i}}{\sum_k e^{s_k}},
\end{equation}
where $U,V\in \mathbb{R}^h$ are learnable parameters, $\tanh(\cdot)$ and $\sigma(\cdot)$ are the hyperbolic tangent and sigmoid, and $\odot$ denotes element-wise multiplication. The slide-level embedding is obtained as follows:
\begin{equation}
	\label{wsifea}
	z_{\mathrm{WSI}}=\sum_{i=1}^M \big(\lambda_\mathrm{SPF}\alpha_i+(1-\lambda_\mathrm{SPF})\beta_i \big)g_i^{\mathrm{AR}},
\end{equation}
where $\lambda_\mathrm{SPF}\in [0,1]$ controlling the trade-off between coverage prior and semantic attention weights.

\noindent\textbf{ROI feature.} For each WSI, we select the adaptive region with the largest SPF weight $\lambda_\mathrm{SPF}\alpha_i+(1-\lambda_\mathrm{SPF})\beta_i$ as the ROI and use its descriptor $g_i^{\mathrm{AR}}$  as the ROI feature. We then evaluate these ROI features on downstream tasks.

\subsection{Model Pretraining Pipeline}
In this paper, we adopt a two-stage training strategy, as illustrated in \cref{framework}. First, we perform self-supervised pretraining on unimodal data by adapting the iBOT algorithm \cite{ibot} for the CARE architecture. Second, we conduct cross-modal contrastive training under the CLIP framework \cite{radford2021learning} using paired WSIs with RNA/protein profiles.

We curate a public multimodal cohort from TCGA \cite{tcga} and GTEx \cite{gtex} for training. The dataset includes 11,463 H$\&$E-stained, formalin-fixed, paraffin-embedded (FFPE) whole-slide images from TCGA and 22,814 normal-tissue slides from GTEx. Within this cohort, we identify 13,289 WSI–RNA pairs and 8,225 WSI–protein pairs.


\subsubsection{Stage I: Unimodal Self-Supervised Pretraining}
We perform unimodal self-supervised pretraining with iBOT, a ViT-based teacher–student framework that predicts masked patch targets with an online prototype vocabulary while enforcing multiview consistency. We adapt iBOT in a backbone-agnostic manner with mechanisms that scale pretraining to gigapixel WSIs.

\noindent\textbf{Balancing batch size and gigapixel WSI.} Gigapixel WSIs yield tens of thousands of patches, constraining training to small batches and weakening self-supervised pretraining. To trade off patch count and batch size, we cluster patch coordinates with DBSCAN \cite{schubert2017dbscan} to partition each slide into sub-WSIs ($\le 360$ patches), converting 34,277 WSIs into 285,710 sub-WSIs and enabling larger effective batch sizes for iBOT pretraining.

\begin{table*}[t]
  \caption{Average ACC (or C-index) by task category across the 33-task benchmark. The best score is in bold, and the second-best is underlined. ``Morph. Class.'' denotes the average performance on the morphological classification task. ``Molecular Class.'' reports the mean result for molecular tasks. ``Molecular $\text{Class.}_V$'' denotes the average result on no-validation subset of molecular tasks. ``Sur. Anal.'' denotes to the average performance on the survival prediction task. Detailed results for the 33 benchmarks are provided in the appendix.}
  \label{sum_result1}
  \centering
  \scalebox{0.96}{
  \begin{tabular}{@{}llllllllll@{}}
    \toprule
    Task  & Head  &\multicolumn{1}{c}{Mean-pool}&\multicolumn{1}{c}{CHIEF} &\multicolumn{1}{c}{PRISM}&\multicolumn{1}{c}{GigaPath}&\multicolumn{1}{c}{TANGLE} &\multicolumn{1}{c}{FEATHER}& \multicolumn{1}{c}{TITAN} & \multicolumn{1}{c}{CARE}\\ \midrule
			
	\multirow{3}{*}{\shortstack{Morph.\\Class.}} &LR& 81.2$\pm$0.1 & 80.2$\pm$0.1 & 75.6$\pm$0.3 & 78.3$\pm$0.1 & 80.9$\pm$0.1 & 81.1$\pm$0.1 & \underline{85.4$\pm$0.1} & \textbf{85.7$\pm$0.0} \\
			& $k$NN & 73.5$\pm$0.1 & 71.4$\pm$0.3 & 68.9$\pm$0.2 & 67.8$\pm$0.1 & 75.2$\pm$0.1 & 72.2$\pm$0.1 & \textbf{83.1$\pm$0.1} & \underline{82.3$\pm$0.1} \\
			&FT& 82.6$\pm$0.1 & 78.0$\pm$0.1 & 81.4$\pm$0.1 & 84.9$\pm$0.1 & 86.1$\pm$0.1 & 86.7$\pm$0.0 & \textbf{87.7$\pm$0.1} & \underline{87.0$\pm$0.1} \\
			\midrule
			\multirow{3}{*}{\shortstack{Molecular\\Class.}}&LR & 67.1$\pm$0.1 & 66.3$\pm$0.1 & 69.1$\pm$0.1 & 66.6$\pm$0.2 & \underline{69.1$\pm$0.1} & 67.6$\pm$0.1 & 67.8$\pm$0.2 & \textbf{69.5$\pm$0.1} \\
			&$k$NN & 64.5$\pm$0.1 & 62.7$\pm$0.1 & 65.2$\pm$0.1 & 62.0$\pm$0.1 & 65.9$\pm$0.1 & 64.6$\pm$0.1 & \underline{66.9$\pm$0.1} & \textbf{68.0$\pm$0.1} \\
			&FT& 70.9$\pm$0.1 & 73.4$\pm$0.2 & 73.2$\pm$0.1 & 73.5$\pm$0.3 & 74.2$\pm$0.2 & \textbf{75.6$\pm$0.1} & 71.7$\pm$0.1 & \underline{74.7$\pm$0.2} \\
			\midrule
			\multirow{2}{*}{\shortstack{{Molecular}\\$\text{Class.}_V$}}&LR&62.7$\pm$1.0&58.8$\pm$0.7&60.5$\pm$0.9&\underline{62.8$\pm$0.9}&60.3$\pm$0.7&59.6$\pm$0.9&  62.5$\pm$0.9&\textbf{64.9$\pm$0.9}\\
			&$k$NN& 55.4$\pm$0.5 & 55.8$\pm$0.5 & 54.4$\pm$0.4 & 55.6$\pm$0.4 & 57.1$\pm$0.5 & 54.9$\pm$0.3 & \underline{57.1$\pm$0.5} & \textbf{57.3$\pm$0.5} \\
			\midrule
			{Sur. Anal.}&linear&48.5$\pm$11.7&	54.3$\pm$10.2&	42.1$\pm$14.1&	\underline{55.7$\pm$9.7}&	49.8$\pm$11.1&	53.9$\pm$12.1&	47.2$\pm$13.0&	\textbf{58.0$\pm$10.4}\\ 		
			
    \bottomrule
  \end{tabular}}
\end{table*}

\noindent\textbf{Frozen feature augmentation.} Since CARE operates on pre-extracted patch features, pixel-level augmentation does not apply. We adopt frozen feature augmentation \cite{bar2024frozen}, perturbing global and local representations in feature space via brightness scaling, contrast adjustment, and posterization-like quantization.

\begin{figure*}
  \centering
  \includegraphics[width=\linewidth]{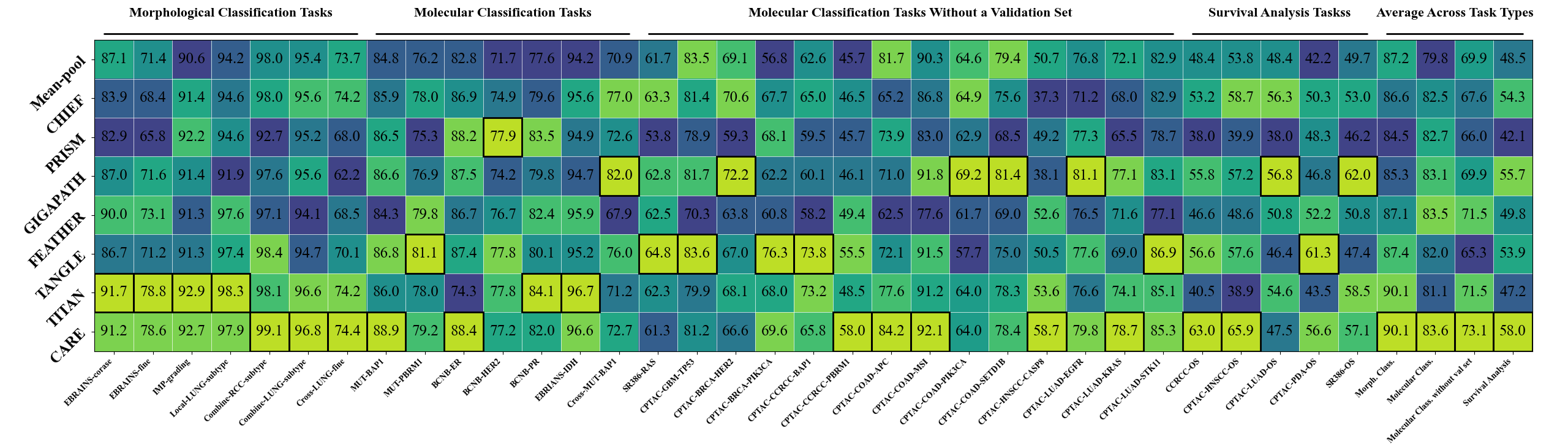}
  \caption{AUROC (or F1) results on 33 downstream tasks under a logistic regression (or a linear layer for survival analysis) setting. The best-performing results are outlined with black boxes.}
  \label{heatmap_auc}
  \vspace{-0.2\baselineskip}
\end{figure*}


\subsubsection{Stage II: Cross-modal Contrastive Pretraining}

\noindent\textbf{RNA-Guided Pretraining.} First, we align paired WSI--RNA data with a CLIP-style objective \cite{radford2021learning}. On the image side, the WSI encoder is initialized from Stage I self-supervision. Guided by the 50 Hallmark gene sets \cite{surv}, we select 3,999 genes and form tokens by summing a learnable RNA-ID embedding with an expression embedding. The token sequence is then encoded by a Transformer, and we mean-pool its outputs to obtain the RNA embedding. The RNA encoder is initialized with scGPT-pretrained weights \cite{cui2023scGPT}. We project both the slide-level and RNA embeddings through lightweight projection heads to a shared embedding space. We then apply $\ell_2$ normalization and optimize a symmetric InfoNCE objective with a learnable temperature.

\noindent\textbf{Protein-Guided Pretraining.} Second, we align WSI–protein pairs, initializing the WSI encoder from the WSI--RNA phase. The protein branch mirrors the RNA pipeline but uses the top-10 proteins by abundance per sample: for each protein we sum a protein-ID embedding with an expression-value embedding, encode the sequence with a Transformer, and take an expression-weighted average of token outputs as the protein embedding. We then apply the same CLIP-style symmetric InfoNCE objective to align protein and WSI embeddings. The protein-ID embedding table is initialized from ESM-2 \cite{esm}, and the remaining parameters are randomly initialized.

Additional pretraining hyperparameters and details are presented in the appendix.




\section{Experiments}

\begin{figure}
  \centering
  \includegraphics[width=\linewidth]{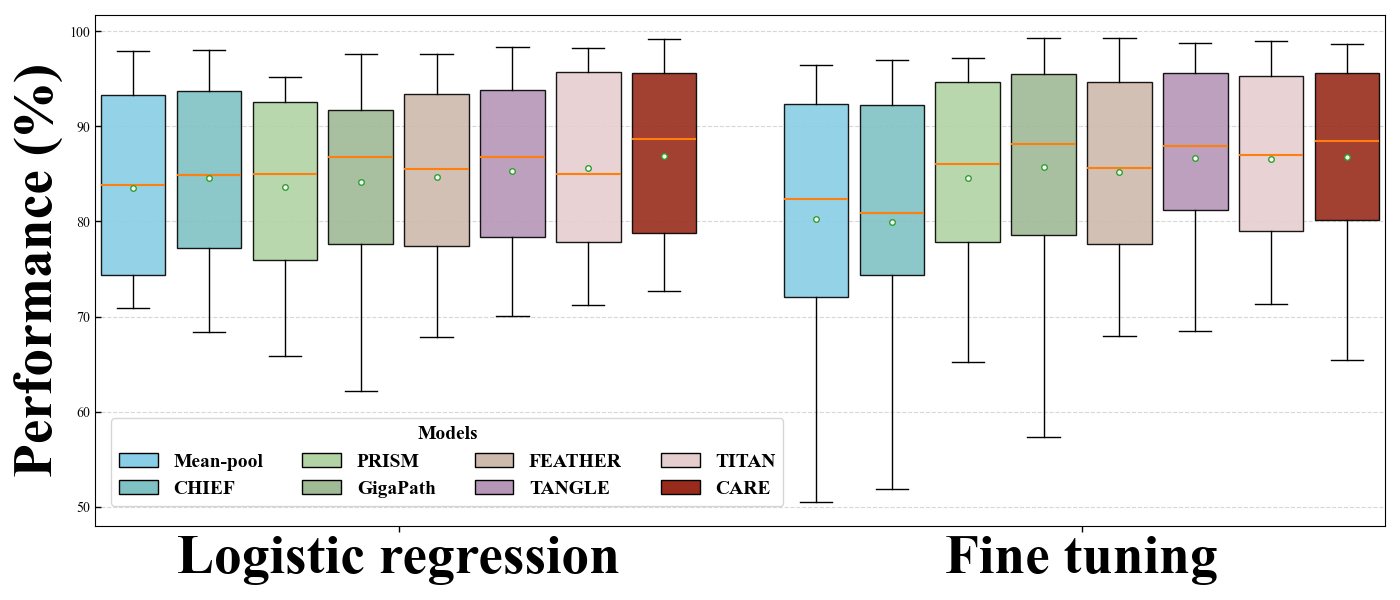}
  \caption{Box plots of experimental results. Each box plot aggregates results from all morphology-classification and molecular-classification tasks. The horizontal line inside each box indicates the median for that model across tasks. The green circle denotes the mean.}
  \label{auc_box}
\end{figure}

\begin{figure*}
  \centering
  \includegraphics[width=\linewidth]{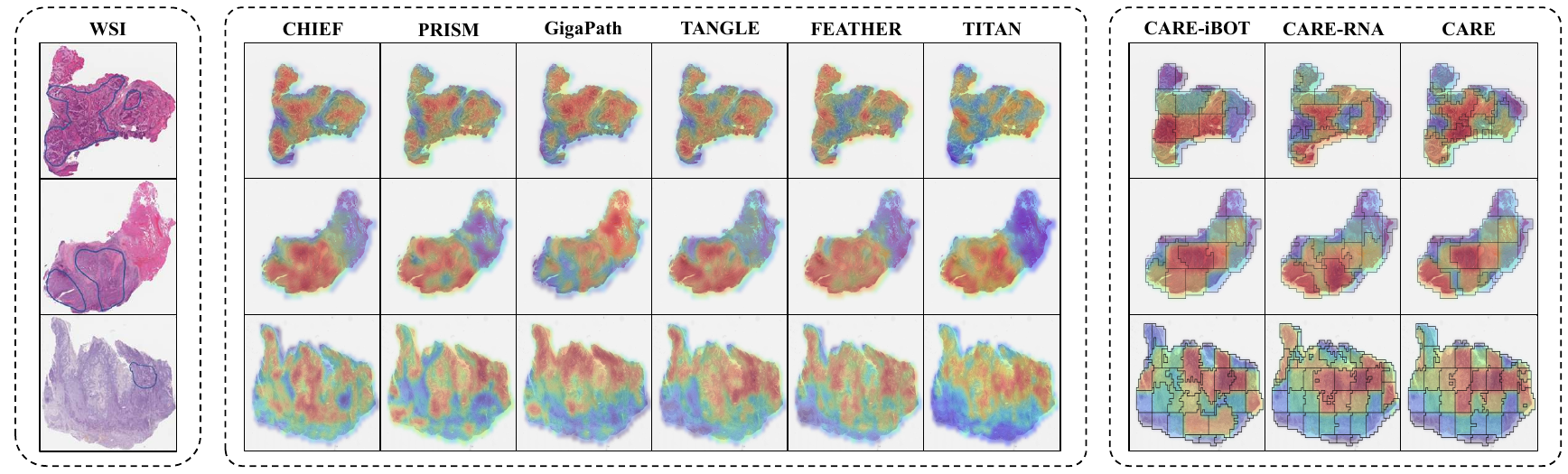}
  \caption{Heatmap visualizations across different foundation models. The WSIs are sourced from the CPTAC-HNSCC dataset. All saliency maps were post-processed with Gaussian blur to smooth region boundaries. Warmer colors (red) indicate stronger model attention.}
  \label{vis}
  \vspace{-0.2\baselineskip}
\end{figure*}

\subsection{Settings}
\noindent\textbf{Datasets}. We evaluate CARE on morphology classification, survival analysis, and molecular prediction tasks across public datasets: CPTAC \cite{cptac}, EBRAINS \cite{roetzer2022digital}, MUT-HET-RCC \cite{acosta2022intratumoral}, IMP \cite{oliveira2021cad}, BCNB \cite{xu2021predicting}, DHMC-LUAD \cite{wei2019pathologist}, DHMC-RCC \cite{zhu2021development}, and SR386 \cite{myles2025surgen}, as well as an institutional cohort (Local-LUNG). Additional details for each dataset are provided in the appendix. 


\noindent\textbf{Implementation Details}. We evaluate two settings: linear probing, fine tuning (FT). For linear probing, we freeze both the patch encoder and the slide-level foundation model and train shallow classifiers, \eg, logistic regression (LR) and $k$NN. In particular, for survival analysis, we use a linear layer to map each WSI’s slide-level features to logits. In the fine-tuning setting, the slide-level foundation model is initialized from pretrained weights. All downstream task evaluations were conducted on a single RTX 4090 GPU\footnote{Code is released at: \url{https://github.com/zdipath/CARE}}. 



\noindent\textbf{Models}. To assess the effectiveness of the CARE framework and pretraining strategy, we benchmark CARE against both a mean-pooling baseline \cite{lu2024visual} and representative slide-level foundation models (CHIEF \cite{chief}, PRISM \cite{prism}, GigaPath \cite{xu2024whole}, FEATHER \cite{feather}, TANGLE \cite{tangle}, TITAN \cite{titan}). The pretraining details of models are presented in the appendix.


\noindent\textbf{Evaluation Metrics}. For morphology classification and molecular prediction, we evaluate binary tasks with balanced accuracy (ACC) and AUC, multiclass tasks with balanced accuracy and macro-F1 (F1), and survival analysis tasks with Harrell’s concordance index (C-index). We run 50 Monte Carlo cross-validation repetitions for datasets without a validation set, and 5 for the rest. Results are reported as mean $\pm$ standard deviation across repetitions.

\subsection{Experiment Results}
We compare state-of-the-art (SOTA) slide-level pathology foundation models under two evaluation regimes: linear probing and fine-tuning. \Cref{sum_result1} reports average balanced accuracy by task type. \Cref{auc_box} shows box plots of AUC (or F1) for tasks with a validation set. Detailed results for the 33 benchmarks are provided in the appendix. Compared with baselines and other foundation models, CARE achieves strong results under both linear probing and fine-tuning. Notably, on the combined RCC subtype-classification task with logistic regression, CARE improves balanced accuracy by 3.8\% points and F1 by 0.7\% points over the next-best model. Specifically, \Cref{heatmap_auc} shows the AUC (or F1) heatmap for the 33 benchmark tasks under a logistic regression setting (or a linear layer for survival analysis). In 28 logistic regression evaluations, CARE achieves SOTA balanced accuracy on 7 tasks and ranks second on 15 tasks. For AUC (or F1 for multiclass tasks, or C-index for survival analysis tasks) in 33 benchmark tasks, it attains SOTA on 12 tasks and the second-best result on 9 tasks. 


\begin{table}
  \caption{Results for ROI feature classification. WSI-feature denotes linear-probing using CARE’s WSI embedding. ROI-feature denotes linear-probing using ROI embeddings of CARE.}
  \label{roi_classification}
  \centering
  \begin{tabular}{@{}l|cc|cc@{}}
    \toprule
    Method     &\multicolumn{2}{c|}{Slide-feature}&\multicolumn{2}{c}{ROI-feature} \\ 
			
			Metric&ACC&AUC &ACC&AUC\\
			\midrule
			CCRCC-BAP1 &58.7&65.8&61.0&69.1\\
			HNSCC-CASP8&51.7&58.7&53.7&61.5\\
			Cross-LUNG-subtype&52.5&	77.0&	56.4& 81.2	\\
    \bottomrule
  \end{tabular}
	\vspace{-0.1\baselineskip}
\end{table}

\begin{table}
  \caption{Ablation study on the ARG. \textit{Not adaptive region} denotes using fixed subregions in place of adaptive regions. All results are reported under the random initialization setting.}
  \label{ablation_result1}
  \centering
  \begin{tabular}{@{}l|cc|cc@{}}
    \toprule
    Method     &\multicolumn{2}{c|}{EBRAINS-fine}&\multicolumn{2}{c}{BCNB-HER2} \\ 
			
			Metric&ACC&F1&ACC&AUC\\
			\midrule
			Not adaptive region&71.6&73.4&61.4&75.0\\
			\midrule
			Not $g^\mathrm{CLS}_i$ feature &71.5&71.9&58.9&73.8\\
			Not  $g^{Q}_i$ feature&71.2&71.7&61.9&74.5\\
			Not  $f_i$ feature&70.1&69.9&61.6&\textbf{75.6}\\
			Not  $f^R_i$ feature&70.8&71.2&59.6&75.4\\
			CARE&\textbf{72.5}&\textbf{74.2}&\textbf{64.8}&{75.4}\\
			\bottomrule
  \end{tabular}
\vspace{-0.1\baselineskip}
\end{table}

Compared with its performance on morphological classification under linear probing, CARE achieves significantly better results on molecular prediction, demonstrating a clear advantage. We argue that the molecule-guided pretraining pathway, together with the lightweight aggregation module SPF, effectively captures biologically relevant tissue areas, thereby improving the performance of molecular prediction. Moreover, CARE remains competitive in morphological classification, thanks to its adaptive region architecture that partitions WSIs into morphologically coherent regions and reduces semantic mixing across tissue types.

\begin{table}
  \caption{Ablation results for CARE hyperparameters. All experiments are reported under the random initialization setting.}
  \label{ablation_parameter}
  \centering
  \begin{tabular}{@{}l|cc|cc@{}}
    \toprule
   &\multicolumn{2}{c|}{EBRAINS-fine}&\multicolumn{2}{c}{BCNB-HER2} \\ 
			
			Metric&ACC&F1&ACC&AUC\\
			\midrule
			Window size $k = 6$&71.2&73.4&59.4&73.3\\
			Window size $k = 8$&\textbf{72.5}&\textbf{74.2}&\textbf{64.8}&\textbf{75.4}\\
			Window size $k = 10$&71.9&73.9&61.9&75.1\\
			\midrule
			$\lambda_\mathrm{RSL} = 0.01$&72.1&73.4&60.2&74.7\\
			$\lambda_\mathrm{RSL} = 0.1$&\textbf{72.5}&\textbf{74.2}&\textbf{64.8}&\textbf{75.4}\\
			$\lambda_\mathrm{RSL} = 0.3$&71.1&72.8&58.6&73.1\\
			\midrule
			$E^\star = 0$&69.1& 70.0&63.9& 74.7\\
			$E^\star = 0.5$&\textbf{72.5}&\textbf{74.2}&\textbf{64.8}&\textbf{75.4}\\
			$E^\star = 2$&69.0& 71.7&59.1 &74.5\\
			\midrule
			$\lambda_\mathrm{SPF} = 0$&70.6&72.5&62.5&74.4\\
			$\lambda_\mathrm{SPF} = 0.5$&\textbf{72.5}&\textbf{74.2}&\textbf{64.8}&\textbf{75.4}\\
			$\lambda_\mathrm{SPF} = 1$&71.2&72.7&60.0&72.7\\

			\bottomrule
  \end{tabular}
\end{table}

To assess the role of ROI in computational pathology, \Cref{roi_classification} compares results using the slide-level feature and the ROI feature. On the CCRCC–BAP1 task, using ROI features with logistic regression yields a performance increase. However, ROI features do not outperform WSI features across all tasks. This phenomenon is consistent with domain knowledge. Some tasks depend on localized signals, whereas others rely on global slide-level context. In the former, CARE’s ROI selection aligns with the task-relevant region and yields gains. In the latter, holistic WSI representations can be equally effective or even superior.

We visualize heatmaps on CPTAC–HNSCC WSIs (\cref{vis}) to examine attended regions. CARE-iBOT denotes iBOT-only pretraining, whereas CARE-RNA uses weights from contrastive WSI–RNA training. All models are shown with pretrained weights only (no downstream fine-tuning). Although heatmaps separate tumor from normal tissue, this contrast does not guarantee clinical relevance. A pathologist annotated (blue contours) regions with marked nuclear atypia and frequent mitoses. CARE focuses on areas matching these annotations, indicating finer-grained diagnostic alignment. Across stages, CARE-iBOT shows suboptimal region partitioning and ROI selection, while CARE-RNA and the full CARE model improve ROI localization.

\subsection{Ablation Studies}
\noindent\textbf{Analysis of ARG}. We ablate adaptive region chunks to test their necessity and the effectiveness of ARG (\cref{ablation_result1}). “No adaptive region” replaces adaptive regions with fixed subregions. All models are trained from random initialization. ARG yields substantial gains, and removing any single component---CLS-aggregated descriptors, query-aggregated descriptors, patch features, or subregion-aware patch features---consistently degrades performance.

\noindent\textbf{Analysis of hyperparameters}. To select CARE’s hyperparameters, we conduct an ablation study (\cref{ablation_parameter}) under random initialization. The best setting is $k=8$, $\lambda_{\mathrm{RSL}}=0.1$, $E^{\star}=0.5$, and $\lambda_{\mathrm{SPF}}=0.5$. Setting $E^{\star}=0$ (fixed regions) degrades performance, confirming the need for adaptive partitioning. $E^{\star}=2$ (biasing to the farthest of the top-3 candidates) hurts. Extreme fusion weights are suboptimal: $\lambda_{\mathrm{SPF}}=0$ collapses SPF to semantic attention, whereas $\lambda_{\mathrm{SPF}}=1$ reduces it to the coverage prior.

\begin{table}
  \caption{Ablation results across pretraining stages. All results are reported under the linear probing setting using logistic regression. CARE-iBOT (less) indicates the version of the model trained via iBOT using less data (24,000 WSIs).}
  \label{ablation_pretraining}
  \centering
  \begin{tabular}{@{}l|cc|cc@{}}
    \toprule
    Method     &\multicolumn{2}{c|}{EBRAINS-fine}&\multicolumn{2}{c}{BCNB-HER2} \\ 
			Metric&ACC&F1&ACC&AUC\\
			\hline
			CARE-ibot (less)&71.5&76.5&56.9&76.8\\
			CARE-ibot&72.2&	77.4&\textbf{60.3} &77.0\\
			CARE-RNA &73.7&	78.5&59.7&76.9\\
			CARE&\textbf{74.0}&\textbf{78.7}&59.8&\textbf{77.2}\\
			\hline
  \end{tabular}
\end{table}

\noindent\textbf{Analysis of pretraining stages}. To assess the effectiveness of pretraining, we evaluate models at different training stages using logistic regression. In \cref{ablation_pretraining}, ``CARE-iBOT (less)'' denotes a variant pretrained on a smaller set of WSIs. We find that scaling self-supervised pretraining data yields a marked boost in initial performance. Subsequently, in the cross-modal stage, leveraging patient-level molecular information, we further improve classification performance by guiding the model toward task-relevant ROIs.
\section{Conclusion}

A key question in CPath is how to partition WSIs into morphology-consistent regions. In this work, we proposed CARE, a slide-level foundation model with an adaptive region generator to address this problem, together with an efficient pretraining pipeline that achieved competitive performance using substantially less data. Our experiments showed that CARE surpassed existing slide-level foundation models and offered a pathology-grounded route to efficient, diagnostically faithful computational pathology.

\section*{Acknowledgements}
This work was partially supported by the National Natural Science Foundation of China (Grant No. 62506291), the National Science and Technology Major Project (Grant No. 2025ZD0544802), the Key Research and Development Program of Shaanxi Province (Grant No. 2024SF-GJHX-32), the Key Research and Development Program of Ningxia Hui Autonomous Region (Grant No. 2023BEG02023), the Noncommunicable Chronic Diseases–National Science and Technology Major Project (Grant No. 2024ZD0527700), the project ``Research on Key Technologies for Full-Chain Intelligent Pathological Diagnosis'' (Grant No. HX202440) of The First Affiliated Hospital of Xi'an Jiaotong University, and GE HealthCare.
{
    \small
    \bibliographystyle{ieeenat_fullname}
    \bibliography{main}

@String(CVPR= {IEEE Conf. Comput. Vis. Pattern Recog.})

@String(CVPR  = {CVPR})

@article{ibot,
	title={ibot: Image bert pre-training with online tokenizer},
	author={Zhou, Jinghao and Wei, Chen and Wang, Huiyu and Shen, Wei and Xie, Cihang and Yuille, Alan and Kong, Tao},
	journal={arXiv preprint arXiv:2111.07832},
	year={2021}
}

@inproceedings{cersovsky2023towards,
  title={Towards hierarchical regional transformer-based multiple instance learning},
  author={Cersovsky, Josef and Mohammadi, Sadegh and Kainmueller, Dagmar and Hoehne, Johannes},
  booktitle={Proceedings of the IEEE/CVF International Conference on Computer Vision},
  pages={3952--3960},
  year={2023}
}

@article{yan2022deep,
  title={Deep contrastive learning based tissue clustering for annotation-free histopathology image analysis},
  author={Yan, Jiangpeng and Chen, Hanbo and Li, Xiu and Yao, Jianhua},
  journal={Computerized Medical Imaging and Graphics},
  volume={97},
  pages={102053},
  year={2022},
  publisher={Elsevier}
}

@article{tcga,
	title={The cancer genome atlas pan-cancer analysis project},
	author={Weinstein, John N and Collisson, Eric A and Mills, Gordon B and Shaw, Kenna R and Ozenberger, Brad A and Ellrott, Kyle and Shmulevich, Ilya and Sander, Chris and Stuart, Joshua M},
	journal={Nature genetics},
	volume={45},
	number={10},
	pages={1113--1120},
	year={2013},
	publisher={Nature Publishing Group}
}

@article{gtex,
	title={The Genotype-Tissue Expression (GTEx) pilot analysis: multitissue gene regulation in humans},
	author={GTEx Consortium and Ardlie, Kristin G and Deluca, David S and Segr{\`e}, Ayellet V and Sullivan, Timothy J and Young, Taylor R and Gelfand, Ellen T and Trowbridge, Casandra A and Maller, Julian B and Tukiainen, Taru and others},
	journal={Science},
	volume={348},
	number={6235},
	pages={648--660},
	year={2015},
	publisher={American Association for the Advancement of Science}
}

@article{cptac,
	title={The CPTAC data portal: a resource for cancer proteomics research},
	author={Edwards, Nathan J and Oberti, Mauricio and Thangudu, Ratna R and Cai, Shuang and McGarvey, Peter B and Jacob, Shine and Madhavan, Subha and Ketchum, Karen A},
	journal={Journal of proteome research},
	volume={14},
	number={6},
	pages={2707--2713},
	year={2015},
	publisher={ACS Publications}
}

@article{roetzer2022digital,
	title={The digital brain tumour atlas, an open histopathology resource},
	author={Roetzer-Pejrimovsky, Thomas and Moser, Anna-Christina and Atli, Baran and Vogel, Clemens Christian and Mercea, Petra A and Prihoda, Romana and Gelpi, Ellen and Haberler, Christine and H{\"o}ftberger, Romana and Hainfellner, Johannes A and others},
	journal={Scientific Data},
	volume={9},
	number={1},
	pages={55},
	year={2022},
	publisher={Nature Publishing Group UK London}
}

@article{acosta2022intratumoral,
	title={Intratumoral resolution of driver gene mutation heterogeneity in renal cancer using deep learning},
	author={Acosta, Paul H and Panwar, Vandana and Jarmale, Vipul and Christie, Alana and Jasti, Jay and Margulis, Vitaly and Rakheja, Dinesh and Cheville, John and Leibovich, Bradley C and Parker, Alexander and others},
	journal={Cancer research},
	volume={82},
	number={15},
	pages={2792--2806},
	year={2022},
	publisher={American Association for Cancer Research}
}

@article{xu2021predicting,
	title={Predicting axillary lymph node metastasis in early breast cancer using deep learning on primary tumor biopsy slides},
	author={Xu, Feng and Zhu, Chuang and Tang, Wenqi and Wang, Ying and Zhang, Yu and Li, Jie and Jiang, Hongchuan and Shi, Zhongyue and Liu, Jun and Jin, Mulan},
	journal={Frontiers in oncology},
	volume={11},
	pages={759007},
	year={2021},
	publisher={Frontiers Media SA}
}

@article{oliveira2021cad,
	title={CAD systems for colorectal cancer from WSI are still not ready for clinical acceptance},
	author={Oliveira, Sara P and Neto, Pedro C and Fraga, Joao and Montezuma, Diana and Monteiro, Ana and Monteiro, Jo{\~a}o and Ribeiro, Liliana and Gon{\c{c}}alves, Sofia and Pinto, Isabel M and Cardoso, Jaime S},
	journal={Scientific Reports},
	volume={11},
	number={1},
	pages={14358},
	year={2021},
	publisher={Nature Publishing Group UK London}
}

@article{wei2019pathologist,
	title={Pathologist-level classification of histologic patterns on resected lung adenocarcinoma slides with deep neural networks},
	author={Wei, Jason W and Tafe, Laura J and Linnik, Yevgeniy A and Vaickus, Louis J and Tomita, Naofumi and Hassanpour, Saeed},
	journal={Scientific reports},
	volume={9},
	number={1},
	pages={3358},
	year={2019},
	publisher={Nature Publishing Group UK London}
}

@inproceedings{radford2021learning,
  title={Learning transferable visual models from natural language supervision},
  author={Radford, Alec and Kim, Jong Wook and Hallacy, Chris and Ramesh, Aditya and Goh, Gabriel and Agarwal, Sandhini and Sastry, Girish and Askell, Amanda and Mishkin, Pamela and Clark, Jack and others},
  booktitle={International conference on machine learning},
  pages={8748--8763},
  year={2021},
  organization={PmLR}
}

@article{myles2025surgen,
	title={SurGen: 1020 H\&E-stained whole-slide images with survival and genetic markers},
	author={Myles, Craig and Um, In Hwa and Marshall, Craig and Harris-Birtill, David and Harrison, David J},
	journal={GigaScience},
	volume={14},
	pages={giaf086},
	year={2025},
	publisher={Oxford University Press}
}

@article{zhu2021development,
	title={Development and evaluation of a deep neural network for histologic classification of renal cell carcinoma on biopsy and surgical resection slides},
	author={Zhu, Mengdan and Ren, Bing and Richards, Ryland and Suriawinata, Matthew and Tomita, Naofumi and Hassanpour, Saeed},
	journal={Scientific reports},
	volume={11},
	number={1},
	pages={7080},
	year={2021},
	publisher={Nature Publishing Group UK London}
}

@article{clam,
	title={Data-efficient and weakly supervised computational pathology on whole-slide images},
	author={Lu, Ming Y and Williamson, Drew FK and Chen, Tiffany Y and Chen, Richard J and Barbieri, Matteo and Mahmood, Faisal},
	journal={Nature biomedical engineering},
	volume={5},
	number={6},
	pages={555--570},
	year={2021},
	publisher={Nature Publishing Group UK London}
}

@inproceedings{li2021dual,
	title={Dual-stream multiple instance learning network for whole slide image classification with self-supervised contrastive learning},
	author={Li, Bin and Li, Yin and Eliceiri, Kevin W},
	booktitle={Proceedings of the IEEE/CVF conference on computer vision and pattern recognition},
	pages={14318--14328},
	year={2021}
}

@inproceedings{zhang2022dtfd,
	title={Dtfd-mil: Double-tier feature distillation multiple instance learning for histopathology whole slide image classification},
	author={Zhang, Hongrun and Meng, Yanda and Zhao, Yitian and Qiao, Yihong and Yang, Xiaoyun and Coupland, Sarah E and Zheng, Yalin},
	booktitle={Proceedings of the IEEE/CVF conference on computer vision and pattern recognition},
	pages={18802--18812},
	year={2022}
}

@article{yao2020whole,
	title={Whole slide images based cancer survival prediction using attention guided deep multiple instance learning networks},
	author={Yao, Jiawen and Zhu, Xinliang and Jonnagaddala, Jitendra and Hawkins, Nicholas and Huang, Junzhou},
	journal={Medical image analysis},
	volume={65},
	pages={101789},
	year={2020},
	publisher={Elsevier}
}

@article{li2023outcome,
	title={Outcome-supervised deep learning on pathologic whole slide images for survival prediction of immunotherapy in patients with non--small cell lung cancer},
	author={Li, Butuo and Yang, Linlin and Zhang, Huan and Li, Haoqian and Jiang, Chao and Yao, Yueyuan and Cheng, Shuping and Zou, Bing and Fan, Bingjie and Dong, Taotao and others},
	journal={Modern Pathology},
	volume={36},
	number={8},
	pages={100208},
	year={2023},
	publisher={Elsevier}
}

@article{he2020integrating,
	title={Integrating spatial gene expression and breast tumour morphology via deep learning},
	author={He, Bryan and Bergenstr{\aa}hle, Ludvig and Stenbeck, Linnea and Abid, Abubakar and Andersson, Alma and Borg, {\AA}ke and Maaskola, Jonas and Lundeberg, Joakim and Zou, James},
	journal={Nature biomedical engineering},
	volume={4},
	number={8},
	pages={827--834},
	year={2020},
	publisher={Nature Publishing Group UK London}
}

@inproceedings{shi2024high,
	title={High-resolution spatial transcriptomics from histology images using histosge},
	author={Shi, Zhiceng and Xue, Shuailin and Zhu, Fangfang and Min, Wenwen},
	booktitle={2024 IEEE International Conference on Bioinformatics and Biomedicine (BIBM)},
	pages={2402--2407},
	year={2024},
	organization={IEEE}
}

@article{huang2025scalable,
	title={Scalable Generation of Spatial Transcriptomics from Histology Images via Whole-Slide Flow Matching},
	author={Huang, Tinglin and Liu, Tianyu and Babadi, Mehrtash and Jin, Wengong and Ying, Rex},
	journal={arXiv preprint arXiv:2506.05361},
	year={2025}
}

@article{wang2023retccl,
	title={RetCCL: Clustering-guided contrastive learning for whole-slide image retrieval},
	author={Wang, Xiyue and Du, Yuexi and Yang, Sen and Zhang, Jun and Wang, Minghui and Zhang, Jing and Yang, Wei and Huang, Junzhou and Han, Xiao},
	journal={Medical image analysis},
	volume={83},
	pages={102645},
	year={2023},
	publisher={Elsevier}
}

@article{huang2023visual,
	title={A visual--language foundation model for pathology image analysis using medical twitter},
	author={Huang, Zhi and Bianchi, Federico and Yuksekgonul, Mert and Montine, Thomas J and Zou, James},
	journal={Nature medicine},
	volume={29},
	number={9},
	pages={2307--2316},
	year={2023},
	publisher={Nature Publishing Group US New York}
}

@article{lu2024visual,
	title={A visual-language foundation model for computational pathology},
	author={Lu, Ming Y and Chen, Bowen and Williamson, Drew FK and Chen, Richard J and Liang, Ivy and Ding, Tong and Jaume, Guillaume and Odintsov, Igor and Le, Long Phi and Gerber, Georg and others},
	journal={Nature medicine},
	volume={30},
	number={3},
	pages={863--874},
	year={2024},
	publisher={Nature Publishing Group US New York}
}

@article{zhang2023biomedclip,
	title={Biomedclip: a multimodal biomedical foundation model pretrained from fifteen million scientific image-text pairs},
	author={Zhang, Sheng and Xu, Yanbo and Usuyama, Naoto and Xu, Hanwen and Bagga, Jaspreet and Tinn, Robert and Preston, Sam and Rao, Rajesh and Wei, Mu and Valluri, Naveen and others},
	journal={arXiv preprint arXiv:2303.00915},
	year={2023}
}

@article{chief,
	title={A pathology foundation model for cancer diagnosis and prognosis prediction},
	author={Wang, Xiyue and Zhao, Junhan and Marostica, Eliana and Yuan, Wei and Jin, Jietian and Zhang, Jiayu and Li, Ruijiang and Tang, Hongping and Wang, Kanran and Li, Yu and others},
	journal={Nature},
	volume={634},
	number={8035},
	pages={970--978},
	year={2024},
	publisher={Nature Publishing Group UK London}
}

@article{xu2024whole,
	title={A whole-slide foundation model for digital pathology from real-world data},
	author={Xu, Hanwen and Usuyama, Naoto and Bagga, Jaspreet and Zhang, Sheng and Rao, Rajesh and Naumann, Tristan and Wong, Cliff and Gero, Zelalem and Gonz{\'a}lez, Javier and Gu, Yu and others},
	journal={Nature},
	volume={630},
	number={8015},
	pages={181--188},
	year={2024},
	publisher={Nature Publishing Group UK London}
}

@article{song2023artificial,
	title={Artificial intelligence for digital and computational pathology},
	author={Song, Andrew H and Jaume, Guillaume and Williamson, Drew FK and Lu, Ming Y and Vaidya, Anurag and Miller, Tiffany R and Mahmood, Faisal},
	journal={Nature Reviews Bioengineering},
	volume={1},
	number={12},
	pages={930--949},
	year={2023},
	publisher={Nature Publishing Group UK London}
}

@inproceedings{shi2024vila,
	title={Vila-mil: Dual-scale vision-language multiple instance learning for whole slide image classification},
	author={Shi, Jiangbo and Li, Chen and Gong, Tieliang and Zheng, Yefeng and Fu, Huazhu},
	booktitle={Proceedings of the IEEE/CVF Conference on Computer Vision and Pattern Recognition},
	pages={11248--11258},
	year={2024}
}

@article{lu2024multimodal,
	title={A multimodal generative AI copilot for human pathology},
	author={Lu, Ming Y and Chen, Bowen and Williamson, Drew FK and Chen, Richard J and Zhao, Melissa and Chow, Aaron K and Ikemura, Kenji and Kim, Ahrong and Pouli, Dimitra and Patel, Ankush and others},
	journal={Nature},
	volume={634},
	number={8033},
	pages={466--473},
	year={2024},
	publisher={Nature Publishing Group UK London}
}

@article{chen2024predicting,
	title={Predicting gastric cancer response to anti-HER2 therapy or anti-HER2 combined immunotherapy based on multi-modal data},
	author={Chen, Zifan and Chen, Yang and Sun, Yu and Tang, Lei and Zhang, Li and Hu, Yajie and He, Meng and Li, Zhiwei and Cheng, Siyuan and Yuan, Jiajia and others},
	journal={Signal transduction and targeted therapy},
	volume={9},
	number={1},
	pages={222},
	year={2024},
	publisher={Nature Publishing Group UK London}
}

@article{wang2022transformer,
	title={Transformer-based unsupervised contrastive learning for histopathological image classification},
	author={Wang, Xiyue and Yang, Sen and Zhang, Jun and Wang, Minghui and Zhang, Jing and Yang, Wei and Huang, Junzhou and Han, Xiao},
	journal={Medical image analysis},
	volume={81},
	pages={102559},
	year={2022},
	publisher={Elsevier}
}

@inproceedings{feather,
	title={Do Multiple Instance Learning Models Transfer?},
	author={Shao, Daniel and Chen, Richard J and Song, Andrew H and Runevic, Joel and Lu, Ming Y and Ding, Tong and Mahmood, Faisal},
	booktitle={International Conference on Machine Learning},
	pages={54219--54238},
	year={2025},
	organization={PMLR}
}

@inproceedings{tangle,
	title={Transcriptomics-guided Slide Representation Learning in Computational Pathology},
	author={Jaume, Guillaume and Oldenburg, Lukas and Vaidya, Anurag Jayant and Chen, Richard J. and Williamson, Drew FK and Peeters, Thomas and Song, Andrew H. and Mahmood, Faisal},
	booktitle={Proceedings of the IEEE/CVF Conference on Computer Vision and Pattern Recognition (CVPR)},
	year={2024}
}

@article{musk,
	title={A vision--language foundation model for precision oncology},
	author={Xiang, Jinxi and Wang, Xiyue and Zhang, Xiaoming and Xi, Yinghua and Eweje, Feyisope and Chen, Yijiang and Li, Yuchen and Bergstrom, Colin and Gopaulchan, Matthew and Kim, Ted and others},
	journal={Nature},
	volume={638},
	number={8051},
	pages={769--778},
	year={2025},
	publisher={Nature Publishing Group UK London}
}

@article{uni, 
	title={Towards a general-purpose foundation model for computational pathology},
	author={Chen, Richard J and Ding, Tong and Lu, Ming Y and Williamson, Drew FK and Jaume, Guillaume and Song, Andrew H and Chen, Bowen and Zhang, Andrew and Shao, Daniel and Shaban, Muhammad and others},
	journal={Nature medicine},
	volume={30},
	number={3},
	pages={850--862},
	year={2024},
	publisher={Nature Publishing Group US New York}
}

@inproceedings{surv,
  title={Modeling dense multimodal interactions between biological pathways and histology for survival prediction},
  author={Jaume, Guillaume and Vaidya, Anurag and Chen, Richard J and Williamson, Drew FK and Liang, Paul Pu and Mahmood, Faisal},
  booktitle={Proceedings of the IEEE/CVF Conference on Computer Vision and Pattern Recognition},
  pages={11579--11590},
  year={2024}
}

@article{esm,
	title={Language models of protein sequences at the scale of evolution enable accurate structure prediction},
	author={Lin, Zeming and Akin, Halil and Rao, Roshan and Hie, Brian and Zhu, Zhongkai and Lu, Wenting and Smetanin, Nikita and dos Santos Costa, Allan and Fazel-Zarandi, Maryam and Sercu, Tom and Candido, Sal and others},
	journal={bioRxiv},
	year={2022},
	publisher={Cold Spring Harbor Laboratory}
}

@inproceedings{abmil,
	title={Attention-based deep multiple instance learning},
	author={Ilse, Maximilian and Tomczak, Jakub and Welling, Max},
	booktitle={Proceedings of the International Conference on Machine Learning},
	pages={2127--2136},
	year={2018},
}

@article{cui2023scgpt,
	title={scGPT: toward building a foundation model for single-cell multi-omics using generative AI},
	author={Cui, Haotian and Wang, Chloe and Maan, Hassaan and Pang, Kuan and Luo, Fengning and Duan, Nan and Wang, Bo},
	journal={Nature methods},
	volume={21},
	number={8},
	pages={1470--1480},
	year={2024},
	publisher={Nature Publishing Group US New York}
}

@inproceedings{hipt,
	title={Scaling vision transformers to gigapixel images via hierarchical self-supervised learning},
	author={Chen, Richard J and Chen, Chengkuan and Li, Yicong and Chen, Tiffany Y and Trister, Andrew D and Krishnan, Rahul G and Mahmood, Faisal},
	booktitle={Proceedings of the IEEE/CVF conference on computer vision and pattern recognition},
	pages={16144--16155},
	year={2022}
}

@inproceedings{rrt,
	title={Feature re-embedding: Towards foundation model-level performance in computational pathology},
	author={Tang, Wenhao and Zhou, Fengtao and Huang, Sheng and Zhu, Xiang and Zhang, Yi and Liu, Bo},
	booktitle={Proceedings of the IEEE/CVF conference on computer vision and pattern recognition},
	pages={11343--11352},
	year={2024}
}

@article{prism,
	title={Prism: A multi-modal generative foundation model for slide-level histopathology},
	author={Shaikovski, George and Casson, Adam and Severson, Kristen and Zimmermann, Eric and Wang, Yi Kan and Kunz, Jeremy D and Retamero, Juan A and Oakley, Gerard and Klimstra, David and Kanan, Christopher and others},
	journal={arXiv preprint arXiv:2405.10254},
	year={2024}
}

@article{titan,
  title={A multimodal whole-slide foundation model for pathology},
  author={Ding, Tong and Wagner, Sophia J and Song, Andrew H and Chen, Richard J and Lu, Ming Y and Zhang, Andrew and Vaidya, Anurag J and Jaume, Guillaume and Shaban, Muhammad and Kim, Ahrong and others},
  journal={Nature Medicine},
  pages={1--13},
  year={2025},
  publisher={Nature Publishing Group}
}

@inproceedings{bar2024frozen,
	title={Frozen feature augmentation for few-shot image classification},
	author={B{\"a}r, Andreas and Houlsby, Neil and Dehghani, Mostafa and Kumar, Manoj},
	booktitle={Proceedings of the IEEE/CVF Conference on Computer Vision and Pattern Recognition},
	pages={16046--16057},
	year={2024}
}

@inproceedings{li2025mico,
  title={MiCo: Multiple Instance Learning with Context-Aware Clustering for Whole Slide Image Analysis},
  author={Li, Junjian and Liu, Jin and Kuang, Hulin and Yue, Hailin and He, Mengshen and Wang, Jianxin},
  booktitle={International Conference on Medical Image Computing and Computer-Assisted Intervention},
  pages={376--385},
  year={2025},
  organization={Springer}
}

@article{li20252,
  title={CA2CL: Cluster-Aware Adversarial Contrastive Learning for Pathological Image Analysis},
  author={Li, Junjian and Kuang, Hulin and Liu, Jin and Yue, Hailin and Wang, Jianxin},
  journal={IEEE Journal of Biomedical and Health Informatics},
  year={2025},
  publisher={IEEE}
}

@article{schubert2017dbscan,
	title={DBSCAN revisited, revisited: why and how you should (still) use DBSCAN},
	author={Schubert, Erich and Sander, J{\"o}rg and Ester, Martin and Kriegel, Hans Peter and Xu, Xiaowei},
	journal={ACM Transactions on Database Systems (TODS)},
	volume={42},
	number={3},
	pages={1--21},
	year={2017},
	publisher={Acm New York, NY, USA}
}

@article{simeoni2025dinov3,
	title={Dinov3},
	author={Sim{\'e}oni, Oriane and Vo, Huy V and Seitzer, Maximilian and Baldassarre, Federico and Oquab, Maxime and Jose, Cijo and Khalidov, Vasil and Szafraniec, Marc and Yi, Seungeun and Ramamonjisoa, Micha{\"e}l and others},
	journal={arXiv preprint arXiv:2508.10104},
	year={2025}
}

@article{yu2022coca,
	title={Coca: Contrastive captioners are image-text foundation models},
	author={Yu, Jiahui and Wang, Zirui and Vasudevan, Vijay and Yeung, Legg and Seyedhosseini, Mojtaba and Wu, Yonghui},
	journal={arXiv preprint arXiv:2205.01917},
	year={2022}
}
}

\clearpage
\setcounter{page}{1}
\maketitlesupplementary
\setcounter{section}{0}
\renewcommand{\thesection}{\Alph{section}}

\renewcommand{\thesubsection}{\thesection.\arabic{subsection}}
\renewcommand{\thesubsubsection}{\thesubsection.\arabic{subsubsection}}
\section{Backbone and Pretraining Details}
Hyperparameter details for CARE are reported in \cref{para_model}. The regional self-attention and regional cross-attention modules use a small number of layers to reduce model complexity. \Cref{para_ibot,para_gene} list key hyperparameters for unimodal and cross-modal pretraining, respectively. Notably, during global and local cropping we avoid fixed rectangular windows. Instead, we sample irregular crops whose areas are proportional to sub-WSI patch counts, better reflecting the inherently irregular spatial distribution of tissue in WSIs. Under a frozen augmentation strategy, we stochastically apply two sequential transforms to each crop, selected from brightness, contrast, and posterize, with sampling governed by preset probabilities. This increases augmentation diversity across crops.

\noindent\textbf{Cross-modal contrastive pretraining order.} We align the slide encoder with RNA embeddings before aligning it with protein embeddings. First, the RNA and protein encoders are highly parameterized, containing a very large number of parameters. Jointly training all pairwise alignments (WSI–RNA–protein) would force smaller batch sizes, weakening representation learning. Second, the RNA encoder can be initialized from scGPT \cite{cui2023scGPT}, whereas the protein encoder has only a protein-ID embedding table for initialization. Consequently, the protein-guided stage primarily provides fine-grained refinement of CARE following the stronger RNA-guided stage.

\section{Effect of Different Pretraining Stages for CARE}
\noindent\textbf{Why iBOT Fits CARE?} iBOT’s $L_\mathrm{MIM}$ objective performs self-distillation between in-view patch tokens. For each masked patch, the student’s representation at that location is aligned with the teacher’s representation at the same location. Unlike a standard ViT, which applies global self-attention to the entire image, CARE restricts attention within an adaptive region. Each patch interacts only with semantically related neighboring tokens, avoiding interference from distant, semantically unrelated regions. This locality has two benefits. First, CARE increases within-region consistency and suppresses cross-region noise. Second, aligning the student and the teacher within the same adaptive region context sharpens and stabilizes the $L_\mathrm{MIM}$ supervisory signal, thereby improving pretraining convergence and downstream performance.
\begin{table}[]
	\centering
    \caption{Hyperparameter Selection for CARE.}
	{
		\begin{tabular}{@{}l|l@{}}
			\toprule
			Hyperparameter &Value\\ 
			\hline
            Regional self-attention depth& 2 \\
            Regional self-attention heads & 8\\
            Regional cross-attention heads& 8 \\
            \hline
            Adaptive region self-attention depth & 5 \\
            Adaptive region self-attention heads & 8 \\
            \hline
            Window size & 8\\
            RSL target & 0.5\\
            $\lambda_\mathrm{SPF}$ & 0.5\\
            Input dimension&768\\
            Output dimension&512\\
			\hline
	\end{tabular}}
	\label{para_model}
\end{table}

\noindent\textbf{Molecularly guided ROI salience.} SPF is a lightweight aggregation module that assigns that assigns region-specific weights to adaptive regions. By inspecting these learned weights, we can localize salient ROIs. A patient’s molecular profile is strongly associated with ROIs in pathology slides. Leveraging this information to guide CARE’s pretraining markedly improves CARE’s ability to partition WSIs into adaptive regions. This step imposes constraints at a functionally more proximal molecular level, thereby improving the consistency of image representations.
\section{Description of Datasets and Tasks}
\begin{table}[]
	\centering
    \caption{Hyperparameter Selection during Unimodal Self-Supervised Pretraining. We use 8×80 GB A100 GPUs for the unimodal pretraining stage.}
	{
		\begin{tabular}{@{}l|l@{}}
			\toprule
			Hyperparameter &Value\\ 
			\hline
            Global crop patch ratio (\% per sub-WSI) & 90\%\\
            Global crop number & 2 \\
            Local crop patch ratio (\% per sub-WSI) & 18\%\\
            Local crop number & 6 \\
            Prediction mask ratio & 0.3\\
            Prediction mask variance & 0.15\\
            Prediction shape & block\\
            Prediction mask start epoch & 0 \\
            Shared head for teacher&True\\
            Shared head for student&False\\
            Dropout &0.1\\
             \hline
             
            \multirow{3}{*}{\shortstack{Frozen augmentation strategy\\for patch feature}} & Brightness \\
            &Contrast \\
            &Posterize \\
            \hline
            Batch size & $8\times 36$\\ 
            Freeze last layer epochs & 4\\
            Automatic per-op mixed precision & FP16\\
            Learning rate schedule& Cosine\\
            Minimum learning rate & 1e-6\\
            Peak learning rate & 1e-4\\
            Number of warmup epochs& 30 \\
            Initial weight decay& 0.04 \\
            Final weight decay& 0.4 \\
            Teacher momentum (start)& 0.998\\
            Teacher momentum (final)& 1\\
            Optimization algorithm&AdamW\\
            Maximum gradient clipping & 2\\
            Epochs & 90\\
			\hline
	\end{tabular}}
	\label{para_ibot}
\end{table}
\noindent\textbf{Datasets.} In this paper, we use nine datasets to construct 33 pathology image analysis benchmarks. The datasets are detailed below.
\begin{itemize}
	\item \textbf{CPTAC \cite{cptac}.} We use data from the Clinical Proteomic Tumor Analysis Consortium (CPTAC), a large-scale National Cancer Institute initiative that provides proteogenomic profiles for multiple tumor types in order to better understand the molecular basis of cancer. Our experiments cover multiple CPTAC cohorts, including glioblastoma multiforme (GBM), breast invasive carcinoma (BRCA), clear cell renal cell carcinoma (CCRCC), colon adenocarcinoma (COAD), head and neck squamous cell carcinoma (HNSCC), lung adenocarcinoma (LUAD), and pancreatic ductal adenocarcinoma (PDA).
	\item \textbf{EBRAINS \cite{roetzer2022digital}.} We evaluate our method on data obtained from EBRAINS, the European Brain Research Infrastructure for neuroscience, computing and brain-related medicine. 
    \item \textbf{MUT-HET-RCC \cite{acosta2022intratumoral}.} The MUT-HET-RCC dataset is a clear cell renal cell carcinoma (ccRCC) cohort to study intratumoral mutation heterogeneity in renal cancer.
    \item \textbf{IMP \cite{oliveira2021cad}.} We use the IMP-CRS 2024 dataset, a large-scale collection of colorectal hematoxylin and eosin (H\&E) WSIs.
    \item \textbf{BCNB \cite{xu2021predicting}.} The Early Breast Cancer Core-Needle Biopsy Whole-Slide Image (BCNB) dataset is a publicly available cohort of 1,058 early breast cancer patients, each associated with a H\&E–stained core-needle biopsy whole-slide image and matched clinical data.
    \item \textbf{DHMC-LUAD \cite{wei2019pathologist}.} The DHMC-LUAD dataset is a publicly available collection of 143 H\&E–stained FFPE WSIs of lung adenocarcinoma (LUAD).
    \item \textbf{DHMC-RCC \cite{zhu2021development}.} DHMC-RCC is composed of 563 publicly accessible H\&E–stained FFPE whole-slide images of renal cell carcinoma (RCC).
    \item \textbf{SR386 \cite{myles2025surgen}.} We use the SR386 cohort from the SurGen dataset, a recently released colorectal cancer digital pathology resource comprising H\&E-stained WSIs with linked clinical, genetic and survival information.
    \item \textbf{Local-LUNG.} We conduct additional validation using an internal lung cancer dataset.
\end{itemize}
\begin{table}[]
	\centering
    \caption{Hyperparameter Selection during molecule-slide cross-modal contrastive pretraining. We use 8×80 GB A100 GPUs for the cross-modal pretraining stage.}
	{
		\begin{tabular}{@{}l|l@{}}
			\toprule
			Hyperparameter &Value\\ 
			\hline
            Gene number & 3999\\
            Dropout &0.2\\
            WSI encoder learning rate &1e-5\\
            Minimum WSI encoder learning rate&1e-6\\
            RNA encoder learning rate&5e-5\\
            Minimum RNA encoder learning rate&1e-5\\
            Other modules learning rate & 1e-4\\    
            Minimum other modules learning rate&1e-5\\
            WSI $\&$ RNA encoder freezing (epochs)&20\\
            Epochs & 150 \\
            Maximum gradient clipping & 2\\
            Batch size & $8\times 6$\\ 
            Gradient accumulation steps &1\\
            Automatic per-op mixed precision & FP16\\
            \hline
            Protein number & 10\\
            Protein selection method (per sample)&top-10\\
            Dropout &0.2\\
            WSI encoder learning rate &1e-5\\
            Minimum WSI encoder learning rate&1e-6\\
            Protein encoder learning rate&1e-4\\
            Minimum protein encoder learning rate&5e-5\\
            Other modules learning rate & 1e-4\\    
            Minimum other modules learning rate&1e-5\\
            WSI $\&$ Protein encoder freezing (epochs)&0\\
            Epochs & 30 \\
            Maximum gradient clipping & 2\\
            Batch size & $8\times 10$\\ 
            Gradient accumulation steps &2\\
            Automatic per-op mixed precision & FP16\\
			\hline
	\end{tabular}}
	\label{para_gene}
\end{table}
\noindent\textbf{Tasks.} Moreover, using either single datasets or combinations of multiple datasets, we construct 33 computational pathology tasks, including morphological classification, molecular classification (gene mutation prediction), and survival prediction. The specific tasks are described as follows:
\begin{table*}[]
	\centering
    \caption{Architectural and pretraining statistics of compared slide-level foundation models. We report only the parameters of the vision-inference module. The superscript * indicates that the model was pre-trained on the TCGA pan-cancer dataset and did not report specific numbers.}
	{
		\begin{tabular}{@{}l|llllll@{}}
			\toprule
			Models     &Patch encoder&Slide aggregator&Output size&FLOPs/G& Para/M & Dataset size\\ 
			\hline
			CHIEF&Ctranspath&Variant of ABMIL&$1\times 768$&0.26&1.05&60{,}530 \\
			PRISM&Virchow&Perceiver&$1\times 1280$&359.36&89.21&587{,}196 \\
			GigaPath&GigaPath&LongNet&$1\times 768$&31.80&86.33&171{,}189 \\
			TANGLE&UNI&ABMIL&$1\times 512$&1.57&5.00&$24{,}000^*$ \\
			FEATHER&CONCH v1.5&ABMIL&$1\times 512$&0.31&0.79&$24{,}000^*$ \\
			TITAN&CONCH v1.5&ViT&$1\times 768$&19.21&47.36& $335,645(423{,}122/182{,}862)$\\
			CARE&CONCH v1.5&CARE&$1\times 512$&15.77&18.79&$34{,}277(13{,}289/8{,}225)$ \\
			\hline
	\end{tabular}}
	\label{flop}
\end{table*}

\begin{table*}[]
	\centering
    \caption{Results under the linear probing (logistic regression) setting for morphologic and molecular classification tasks. The best score is in bold, and the second-best is underlined.}
	{
		\begin{tabular}{@{}llllllllll@{}}
			\toprule
			Task  & Metric  &\multicolumn{1}{c}{Mean-pool}&\multicolumn{1}{c}{CHIEF} &\multicolumn{1}{c}{PRISM}&\multicolumn{1}{c}{GigaPath}&\multicolumn{1}{c}{TANGLE} &\multicolumn{1}{c}{FEATHER}& \multicolumn{1}{c}{TITAN} & \multicolumn{1}{c}{CARE}\\ \midrule
			\multirow{2}{*}{Task 1} & ACC & 77.8$\pm$0.1 & 73.0$\pm$0.1 & 72.2$\pm$0.2 & 75.8$\pm$0.1 & 76.4$\pm$0.1 & 82.3$\pm$0.1 & \textbf{87.1$\pm$0.0} & \underline{85.1$\pm$0.1} \\
			& F1 & 87.1$\pm$0.0 & 83.9$\pm$0.0 & 82.9$\pm$0.0 & 87.0$\pm$0.0 & 86.7$\pm$0.0 & 90.0$\pm$0.0 & \textbf{91.7$\pm$0.0} & \underline{91.2$\pm$0.0} \\

            \multirow{2}{*}{Task 2} & ACC & 65.8$\pm$0.0 & 60.6$\pm$0.1 & 59.5$\pm$0.0 & 64.7$\pm$0.1 & 64.5$\pm$0.0 & 68.2$\pm$0.0 & \textbf{74.8$\pm$0.0} & \underline{74.0$\pm$0.1} \\
			& F1 & 71.4$\pm$0.0 & 68.4$\pm$0.1 & 65.8$\pm$0.0 & 71.6$\pm$0.1 & 71.2$\pm$0.0 & 73.1$\pm$0.0 & \textbf{78.8$\pm$0.0} & \underline{78.7$\pm$0.0} \\
			\multirow{2}{*}{Task 3} & ACC & 90.0$\pm$0.0 & 91.4$\pm$0.0 & 91.9$\pm$0.0 & 91.1$\pm$0.0 & 91.0$\pm$0.0 & 91.0$\pm$0.0 & \textbf{92.8$\pm$0.0} & \underline{92.6$\pm$0.0} \\
			& F1 & 90.6$\pm$0.0 & 91.4$\pm$0.0 & 92.2$\pm$0.0 & 91.3$\pm$0.0 & 91.3$\pm$0.0 & 91.3$\pm$0.0 & \textbf{92.9$\pm$0.0} & \underline{92.7$\pm$0.0} \\
			\multirow{2}{*}{Task 4} & ACC & 89.1$\pm$0.5 & 89.6$\pm$0.1 & 90.4$\pm$0.1 & 84.4$\pm$0.4 & 94.2$\pm$0.1 & 95.3$\pm$0.1 & \textbf{96.5$\pm$0.0} & \underline{96.2$\pm$0.0} \\
			& F1 & 94.2$\pm$0.1 & 94.5$\pm$0.0 & 94.6$\pm$0.0 & 91.9$\pm$0.1 & 97.4$\pm$0.0 & 97.6$\pm$0.0 & \textbf{98.3$\pm$0.0} & \underline{97.9$\pm$0.0} \\
			\multirow{2}{*}{Task 5} & ACC & \underline{93.8$\pm$0.1} & 93.5$\pm$0.2 & 70.7$\pm$1.6 & 89.7$\pm$0.2 & 93.3$\pm$0.1 & 88.9$\pm$0.3 & 93.3$\pm$0.4 & \textbf{97.6$\pm$0.1} \\
			& F1 & 98.0$\pm$0.0 & 98.0$\pm$0.0 & 92.7$\pm$0.0 & 97.6$\pm$0.0 & \underline{98.4$\pm$0.0} & 97.1$\pm$0.0 & 98.1$\pm$0.0 & \textbf{99.1$\pm$0.0} \\
			\multirow{2}{*}{Task 6} & ACC & 88.1$\pm$0.0 & 87.6$\pm$0.0 & 86.2$\pm$0.0 & 88.4$\pm$0.1 & 86.3$\pm$0.1 & 85.5$\pm$0.0 & \textbf{89.2$\pm$0.0} & \underline{89.0$\pm$0.0} \\
			& AUC & 95.4$\pm$0.0 & 95.6$\pm$0.0 & 95.2$\pm$0.0 & 95.6$\pm$0.0 & 94.7$\pm$0.0 & 94.1$\pm$0.0 & \underline{96.6$\pm$0.0} & \textbf{96.8$\pm$0.0} \\
			\multirow{2}{*}{Task 7} & ACC & 63.6$\pm$0.1 & \textbf{65.7$\pm$0.1} & 58.3$\pm$0.1 & 54.2$\pm$0.1 & 60.3$\pm$0.0 & 56.2$\pm$0.3 & 63.8$\pm$0.1 & \underline{65.5$\pm$0.0} \\
			& F1 & 73.7$\pm$0.0 & 74.2$\pm$0.0 & 68.0$\pm$0.1 & 62.2$\pm$0.1 & 70.1$\pm$0.0 & 68.5$\pm$0.3 & \underline{74.2$\pm$0.0} & \textbf{74.4$\pm$0.0} \\
			\multirow{2}{*}{Task 8} & ACC & 61.0$\pm$0.0 & 57.2$\pm$0.1 & 57.5$\pm$0.1 & \underline{61.4$\pm$0.1} & \textbf{63.6$\pm$0.2} & 54.7$\pm$0.0 & 59.8$\pm$0.1 & \underline{61.4$\pm$0.1} \\
			& AUC & 84.8$\pm$0.1 & 85.9$\pm$0.0 & 86.5$\pm$0.0 & 86.6$\pm$0.0 & \underline{86.8$\pm$0.0} & 84.3$\pm$0.1 & 86.0$\pm$0.0 & \textbf{88.9$\pm$0.0} \\
			\multirow{2}{*}{Task 9} & ACC & 69.4$\pm$0.1 & 70.2$\pm$0.1 & 68.7$\pm$0.2 & 69.9$\pm$0.1 & \textbf{73.5$\pm$0.1} & \textbf{73.5$\pm$0.1} & 70.3$\pm$0.1 & \underline{71.8$\pm$0.2} \\
			& AUC & 76.2$\pm$0.1 & 78.0$\pm$0.1 & 75.3$\pm$0.1 & 76.9$\pm$0.1 & \textbf{81.1$\pm$0.1} & \underline{79.8$\pm$0.1} & 78.0$\pm$0.1 & 79.2$\pm$0.2 \\

			\multirow{2}{*}{Task 10} & ACC & 70.2$\pm$0.1 & 72.9$\pm$0.6 & \textbf{74.5$\pm$0.2} & 71.1$\pm$0.6 & 71.6$\pm$0.1 & 74.0$\pm$0.2 & 62.9$\pm$1.0 & \underline{74.4$\pm$0.1} \\
			& AUC & 82.8$\pm$0.2 & 86.9$\pm$0.2 & \underline{88.2$\pm$0.1} & 87.5$\pm$0.2 & 87.4$\pm$0.1 & 86.7$\pm$0.2 & 74.3$\pm$1.0 & \textbf{88.4$\pm$0.1} \\
			\multirow{2}{*}{Task 11} & ACC & 57.7$\pm$0.1 & 57.7$\pm$0.0 & \textbf{64.5$\pm$0.1} & 57.8$\pm$0.1 & \underline{61.7$\pm$0.1} & 59.0$\pm$0.1 & 60.7$\pm$0.1 & 59.8$\pm$0.1 \\
			& AUC & 71.7$\pm$0.1 & 74.9$\pm$0.0 & \textbf{77.9$\pm$0.0} & 74.2$\pm$0.0 & \underline{77.8$\pm$0.0} & 76.7$\pm$0.1 & 77.8$\pm$0.1 & 77.2$\pm$0.1 \\
			\multirow{2}{*}{Task 12} & ACC & 65.1$\pm$0.1 & 66.6$\pm$0.0 & \textbf{72.1$\pm$0.0} & 66.5$\pm$0.1 & 66.8$\pm$0.1 & 71.7$\pm$0.0 & \underline{71.8$\pm$0.0} & 70.5$\pm$0.1 \\
			& AUC & 77.6$\pm$0.1 & 79.6$\pm$0.1 & \underline{83.5$\pm$0.1} & 79.8$\pm$0.1 & 80.1$\pm$0.1 & 82.4$\pm$0.1 & \textbf{84.1$\pm$0.0} & 82.0$\pm$0.1 \\
			
			\multirow{2}{*}{Task 13} & ACC & 87.8$\pm$0.1 & 88.9$\pm$0.0 & 88.7$\pm$0.0 & 88.2$\pm$0.0 & 89.0$\pm$0.1 & 88.5$\pm$0.1 & \underline{91.4$\pm$0.0} & \textbf{91.5$\pm$0.0} \\
			& AUC & 94.2$\pm$0.0 & 95.6$\pm$0.0 & 94.9$\pm$0.0 & 94.7$\pm$0.0 & 95.2$\pm$0.0 & 95.9$\pm$0.0 & \textbf{96.7$\pm$0.0} & \underline{96.6$\pm$0.0} \\
			\multirow{2}{*}{Task 14} & ACC & \textbf{58.3$\pm$0.1} & 50.8$\pm$0.0 & \underline{57.6$\pm$0.1} & 51.6$\pm$0.2 & 57.6$\pm$0.2 & 52.0$\pm$0.0 & 57.4$\pm$0.2 & 56.9$\pm$0.1 \\
			& AUC & 70.9$\pm$0.1 & \underline{77.0$\pm$0.0} & 72.6$\pm$0.0 & \textbf{82.0$\pm$0.0} & 76.0$\pm$0.1 & 67.9$\pm$0.2 & 71.2$\pm$0.0 & 72.7$\pm$0.1 \\
			 \bottomrule
	\end{tabular}}
	\label{result_logistic}
\end{table*}
\begin{table*}[t]
	\centering
    \caption{Results on linear probing (logistic regression) setting for molecular classification tasks without a validation set. The best score is in bold, and the second-best is underlined.}
	{
		\begin{tabular}{@{}llllllllll@{}}
			\toprule
			Task  & Metric  &\multicolumn{1}{c}{Mean-pool}&\multicolumn{1}{c}{CHIEF} &\multicolumn{1}{c}{PRISM}&\multicolumn{1}{c}{GigaPath}&\multicolumn{1}{c}{TANGLE} &\multicolumn{1}{c}{FEATHER}& \multicolumn{1}{c}{TITAN} & \multicolumn{1}{c}{CARE}\\ \midrule

			\multirow{2}{*}{Task 15} & ACC & 59.1$\pm$0.3 & 56.2$\pm$0.2 & 51.9$\pm$0.2 & 57.1$\pm$0.2 & \underline{58.1$\pm$0.2} & \textbf{59.8$\pm$0.2} & 57.9$\pm$0.2 & 57.2$\pm$0.3 \\
			& AUC & 61.7$\pm$0.4 & \underline{63.3$\pm$0.2} & 53.9$\pm$0.3 & 62.8$\pm$0.2 & \textbf{64.8$\pm$0.3} & 62.5$\pm$0.3 & 62.3$\pm$0.4 & 61.3$\pm$0.2 \\
			
			\multirow{2}{*}{Task 16} & ACC & \textbf{75.8$\pm$0.7} & 72.8$\pm$0.8 & 67.9$\pm$0.6 & 71.8$\pm$1.0 & 75.0$\pm$0.8 & 62.7$\pm$1.2 & 72.9$\pm$0.8 & \underline{75.3$\pm$0.8} \\
			& AUC & \underline{83.5$\pm$0.9} & 81.4$\pm$1.1 & 78.9$\pm$0.6 & 81.7$\pm$0.9 & \textbf{83.6$\pm$0.9} & 70.3$\pm$1.4 & 79.9$\pm$0.9 & 81.2$\pm$0.8 \\
			\multirow{2}{*}{Task 17} & ACC & 63.1$\pm$1.5 & \underline{64.1$\pm$1.4} & 54.8$\pm$1.2 & 62.0$\pm$0.9 & 55.1$\pm$0.7 & 59.6$\pm$1.2 & 62.8$\pm$1.0 & \textbf{64.5$\pm$0.8} \\
			& AUC & 69.1$\pm$1.5 & \underline{70.6$\pm$1.8} & 59.3$\pm$1.5 & \textbf{72.2$\pm$1.4} & 67.0$\pm$1.2 & 63.9$\pm$1.2 & 68.1$\pm$1.1 & 66.6$\pm$1.1 \\
			\multirow{2}{*}{Task 18} & ACC & 55.6$\pm$1.1 & 57.8$\pm$1.0 & 61.9$\pm$0.8 & 61.9$\pm$0.6 & \textbf{67.9$\pm$1.0} & 56.8$\pm$0.8 & 58.9$\pm$1.1 & \underline{62.4$\pm$1.1} \\
			& AUC & 56.8$\pm$1.3 & 67.7$\pm$1.2 & 68.1$\pm$1.5 & 62.2$\pm$0.9 & \textbf{76.3$\pm$1.8} & 60.8$\pm$1.6 & 68.0$\pm$1.1 & \underline{69.6$\pm$1.5} \\
			\multirow{2}{*}{Task 19} & ACC & 57.5$\pm$0.8 & 56.4$\pm$0.7 & 54.0$\pm$1.1 & 55.2$\pm$0.7 & 56.4$\pm$0.8 & 58.1$\pm$1.6 & \textbf{61.2$\pm$1.4} & \underline{58.7$\pm$1.0} \\
			& AUC & 62.6$\pm$2.9 & 65.0$\pm$2.3 & 59.5$\pm$2.8 & 60.1$\pm$2.8 & \textbf{73.8$\pm$1.6} & 58.2$\pm$3.7 & \underline{73.2$\pm$2.1} & 65.8$\pm$2.6 \\
			\multirow{2}{*}{Task 20} & ACC & 47.2$\pm$0.9 & 47.2$\pm$0.7 & 47.3$\pm$0.8 & 46.6$\pm$0.6 & \underline{53.4$\pm$0.7} & 50.5$\pm$0.9 & 49.5$\pm$0.7 & \textbf{56.0$\pm$0.7} \\
			& AUC & 45.7$\pm$1.3 & 46.5$\pm$1.3 & 45.7$\pm$1.1 & 46.1$\pm$1.4 & \underline{55.5$\pm$1.2} & 49.4$\pm$1.4 & 48.5$\pm$1.3 & \textbf{58.0$\pm$1.0} \\
			\multirow{2}{*}{Task 21} & ACC & \textbf{66.8$\pm$1.8} & 54.1$\pm$0.7 & 59.9$\pm$2.0 & 63.4$\pm$1.6 & 58.7$\pm$0.6 & 57.5$\pm$1.2 & 60.8$\pm$1.2 & \underline{66.6$\pm$1.4} \\
			& AUC & \underline{81.7$\pm$1.2} & 65.2$\pm$1.5 & 73.9$\pm$1.6 & 71.0$\pm$1.4 & 72.1$\pm$1.4 & 62.5$\pm$1.8 & 77.6$\pm$1.1 & \textbf{84.2$\pm$0.5} \\
			\multirow{2}{*}{Task 22} & ACC & \underline{77.4$\pm$1.1} & 70.4$\pm$0.9 & 72.5$\pm$0.7 & 77.1$\pm$1.1 & 69.6$\pm$1.2 & 66.9$\pm$1.0 & \textbf{79.6$\pm$1.1} & 75.7$\pm$0.9 \\
			& AUC & 90.3$\pm$0.4 & 86.8$\pm$0.8 & 83.0$\pm$0.7 & \underline{91.8$\pm$0.6} & 91.5$\pm$0.4 & 77.6$\pm$1.3 & 91.2$\pm$0.5 & \textbf{92.1$\pm$0.3} \\
			\multirow{2}{*}{Task 23} & ACC & 58.6$\pm$1.1 & 53.6$\pm$0.5 & 57.0$\pm$0.9 & \textbf{65.2$\pm$1.3} & 50.6$\pm$0.4 & 54.9$\pm$1.0 & 56.8$\pm$0.9 & \underline{60.0$\pm$1.1} \\
			& AUC & 64.6$\pm$1.7 & \underline{64.9$\pm$1.4} & 62.9$\pm$0.9 & \textbf{69.2$\pm$1.9} & 57.7$\pm$1.2 & 61.7$\pm$1.4 & 64.0$\pm$0.9 & 64.0$\pm$1.2 \\
			\multirow{2}{*}{Task 24} & ACC & 62.3$\pm$1.7 & 54.4$\pm$0.7 & 63.8$\pm$2.0 & \underline{64.3$\pm$2.0} & 58.6$\pm$1.3 & 55.9$\pm$1.0 & 61.0$\pm$1.8 & \textbf{65.3$\pm$1.8} \\
			& AUC & \underline{79.4$\pm$1.9} & 75.7$\pm$3.3 & 68.5$\pm$4.0 & \textbf{81.4$\pm$1.7} & 75.0$\pm$2.4 & 69.0$\pm$2.5 & 78.3$\pm$1.4 & 78.4$\pm$1.7 \\
			\multirow{2}{*}{Task 25} & ACC & 49.3$\pm$0.7 & 47.5$\pm$0.1 & \textbf{54.6$\pm$1.2} & 46.8$\pm$0.1 & 47.9$\pm$0.1 & \underline{51.8$\pm$0.6} & 49.0$\pm$0.3 & 51.7$\pm$1.0 \\
			& AUC & 50.7$\pm$3.4 & 37.3$\pm$4.5 & 49.2$\pm$5.9 & 38.1$\pm$3.5 & 50.5$\pm$2.4 & 52.6$\pm$5.8 & \underline{53.6$\pm$4.4} & \textbf{58.7$\pm$4.8} \\
			\multirow{2}{*}{Task 26} & ACC & 70.0$\pm$0.5 & 65.4$\pm$0.7 & 69.3$\pm$0.5 & \textbf{74.2$\pm$0.7} & 68.4$\pm$0.8 & 69.5$\pm$0.5 & 68.9$\pm$0.6 & \underline{72.2$\pm$0.5} \\
			& AUC & 76.8$\pm$0.7 & 71.2$\pm$1.0 & 77.3$\pm$0.6 & \textbf{81.1$\pm$0.6} & 77.6$\pm$0.8 & 76.5$\pm$0.6 & 76.6$\pm$0.6 & \underline{79.8$\pm$0.5} \\
			\multirow{2}{*}{Task 27} & ACC & 65.9$\pm$0.7 & 59.9$\pm$0.9 & 60.5$\pm$0.3 & \underline{69.1$\pm$1.0} & 61.2$\pm$0.7 & 63.8$\pm$0.9 & 65.6$\pm$0.5 & \textbf{69.5$\pm$0.7} \\
			& AUC & 72.1$\pm$0.6 & 68.0$\pm$1.2 & 65.5$\pm$0.5 & \underline{77.1$\pm$1.0} & 69.0$\pm$1.0 & 71.6$\pm$1.0 & 74.1$\pm$0.6 & \textbf{78.7$\pm$0.7} \\
			\multirow{2}{*}{Task 28} & ACC & 69.6$\pm$0.8 & 63.2$\pm$1.0 & \underline{71.9$\pm$1.0} & 64.9$\pm$1.1 & 63.6$\pm$0.9 & 66.6$\pm$1.1 & 70.4$\pm$1.2 & \textbf{73.1$\pm$1.2} \\
			& AUC & 82.9$\pm$1.0 & 82.9$\pm$1.0 & 78.7$\pm$1.7 & 83.1$\pm$0.6 & \textbf{86.9$\pm$0.5} & 77.1$\pm$1.8 & 85.1$\pm$0.6 & \underline{85.3$\pm$0.6} \\

			 \bottomrule
	\end{tabular}}

	\label{result_logistic_cptac}
\end{table*}

\begin{table*}[]
	\centering
    \caption{Results on linear probing ($k$NN parameter) setting for morphologic and molecular classification tasks. The best score is in bold, and the second-best is underlined.}
	{
		\begin{tabular}{@{}llllllllll@{}}
			\toprule
			Task  & Metric  &\multicolumn{1}{c}{Mean-pool}&\multicolumn{1}{c}{CHIEF} &\multicolumn{1}{c}{PRISM}&\multicolumn{1}{c}{GigaPath}&\multicolumn{1}{c}{TANGLE} &\multicolumn{1}{c}{FEATHER}& \multicolumn{1}{c}{TITAN} & \multicolumn{1}{c}{CARE}\\ \midrule

				\multirow{2}{*}{Task 1} & ACC & 71.4$\pm$0.0 & 55.8$\pm$0.1 & 57.8$\pm$0.1 & 58.7$\pm$0.1 & 66.1$\pm$0.2 & 70.8$\pm$0.1 & \textbf{83.5$\pm$0.1} & \underline{80.9$\pm$0.0} \\
			& F1 & 83.8$\pm$0.0 & 72.4$\pm$0.0 & 73.1$\pm$0.0 & 75.4$\pm$0.0 & 80.7$\pm$0.0 & 84.0$\pm$0.0 & \textbf{90.7$\pm$0.0} & \underline{89.4$\pm$0.0} \\
			\multirow{2}{*}{Task 2} & ACC & 56.6$\pm$0.0 & 44.8$\pm$0.1 & 48.0$\pm$0.0 & 46.4$\pm$0.0 & 54.1$\pm$0.0 & 56.9$\pm$0.0 & \textbf{70.5$\pm$0.0} & \underline{66.2$\pm$0.0} \\
			& F1 & 63.3$\pm$0.0 & 53.3$\pm$0.1 & 54.5$\pm$0.0 & 55.4$\pm$0.0 & 62.8$\pm$0.0 & 65.2$\pm$0.0 & \textbf{75.8$\pm$0.0} & \underline{72.0$\pm$0.0} \\
			
			\multirow{2}{*}{Task 3} & ACC & 84.5$\pm$0.0 & \underline{90.3$\pm$0.0} & 88.6$\pm$0.0 & 84.5$\pm$0.0 & 87.8$\pm$0.0 & 81.1$\pm$0.0 & \textbf{90.5$\pm$0.0} & 89.7$\pm$0.0 \\
			& F1 & 86.7$\pm$0.0 & 90.3$\pm$0.0 & 89.4$\pm$0.0 & 86.6$\pm$0.0 & 89.1$\pm$0.0 & 85.6$\pm$0.0 & \textbf{91.3$\pm$0.0} & \underline{90.4$\pm$0.0} \\

			\multirow{2}{*}{Task 4} & ACC & 77.1$\pm$0.6 & 85.9$\pm$0.3 & 90.0$\pm$0.3 & 73.4$\pm$0.2 & 90.8$\pm$0.1 & 78.0$\pm$0.4 & \underline{97.1$\pm$0.1} & \textbf{98.0$\pm$0.1} \\
			& F1 & 88.3$\pm$0.1 & 93.7$\pm$0.1 & 94.7$\pm$0.1 & 86.7$\pm$0.0 & 96.0$\pm$0.0 & 89.5$\pm$0.1 & \underline{98.1$\pm$0.0} & \textbf{98.6$\pm$0.0} \\
			\multirow{2}{*}{Task 5} & ACC & 84.5$\pm$0.1 & 85.0$\pm$1.1 & 55.7$\pm$0.7 & 81.5$\pm$0.1 & 84.9$\pm$0.1 & 86.9$\pm$0.4 & \underline{92.9$\pm$0.3} & \textbf{96.1$\pm$0.1} \\
			& F1 & 96.4$\pm$0.0 & 96.1$\pm$0.0 & 89.6$\pm$0.0 & 96.1$\pm$0.0 & 96.4$\pm$0.0 & 97.3$\pm$0.0 & \underline{98.0$\pm$0.0} & \textbf{98.1$\pm$0.0} \\
			
			\multirow{2}{*}{Task 6} & ACC & 86.6$\pm$0.1 & 84.7$\pm$0.1 & 84.8$\pm$0.0 & 85.3$\pm$0.1 & 87.7$\pm$0.0 & 85.2$\pm$0.0 & \underline{89.0$\pm$0.0} & \textbf{89.3$\pm$0.1} \\
			& AUC & 95.3$\pm$0.0 & 92.9$\pm$0.1 & 93.6$\pm$0.0 & 92.9$\pm$0.0 & 95.1$\pm$0.0 & 93.6$\pm$0.0 & \textbf{96.2$\pm$0.0} & \underline{96.1$\pm$0.0} \\
			\multirow{2}{*}{Task 7} & ACC & 53.9$\pm$0.0 & 53.3$\pm$0.3 & \underline{57.2$\pm$0.1} & 45.2$\pm$0.1 & 55.2$\pm$0.2 & 46.3$\pm$0.0 & \textbf{58.0$\pm$0.2} & 56.0$\pm$0.1 \\
			& F1 & 61.3$\pm$0.1 & 63.6$\pm$0.4 & 64.0$\pm$0.1 & 52.8$\pm$0.1 & 65.4$\pm$0.1 & 57.5$\pm$0.1 & \textbf{66.9$\pm$0.2} & \underline{66.7$\pm$0.0} \\

			\multirow{2}{*}{Task 8} & ACC & 60.7$\pm$0.1 & 56.8$\pm$0.1 & 57.1$\pm$0.1 & 57.5$\pm$0.1 & 58.6$\pm$0.1 & 56.4$\pm$0.3 & \textbf{62.8$\pm$0.2} & \underline{61.5$\pm$0.1} \\
			& AUC & 71.8$\pm$0.3 & 75.4$\pm$0.1 & 70.6$\pm$0.2 & 73.0$\pm$0.2 & 72.8$\pm$0.2 & 73.4$\pm$0.1 & \underline{75.6$\pm$0.2} & \textbf{78.6$\pm$0.3} \\
			\multirow{2}{*}{Task 9} & ACC & 63.3$\pm$0.1 & 67.6$\pm$0.1 & 60.7$\pm$0.1 & 63.9$\pm$0.1 & \textbf{68.8$\pm$0.1} & 65.9$\pm$0.1 & 66.5$\pm$0.1 & \underline{67.9$\pm$0.2} \\
			& AUC & 68.8$\pm$0.1 & 72.2$\pm$0.1 & 66.5$\pm$0.3 & 70.4$\pm$0.1 & \textbf{75.9$\pm$0.1} & 71.6$\pm$0.2 & 73.5$\pm$0.2 & \underline{74.9$\pm$0.3} \\
			\multirow{2}{*}{Task 10} & ACC & 60.9$\pm$0.2 & 61.2$\pm$0.1 & \textbf{68.2$\pm$0.1} & 57.9$\pm$0.1 & 63.0$\pm$0.1 & 60.8$\pm$0.0 & \underline{67.0$\pm$0.2} & 65.5$\pm$0.1 \\
			& AUC & 72.1$\pm$0.1 & 70.9$\pm$0.6 & \textbf{84.7$\pm$0.1} & 72.0$\pm$0.9 & 75.8$\pm$0.2 & 74.7$\pm$0.1 & 77.3$\pm$0.3 & \underline{82.5$\pm$0.2} \\
			\multirow{2}{*}{Task 11} & ACC & 52.7$\pm$0.0 & 55.5$\pm$0.1 & \underline{60.5$\pm$0.1} & 53.1$\pm$0.0 & 57.4$\pm$0.1 & 56.7$\pm$0.1 & \textbf{64.7$\pm$0.0} & 60.0$\pm$0.0 \\
			& AUC & 65.7$\pm$0.2 & 70.5$\pm$0.1 & \underline{76.8$\pm$0.0} & 64.9$\pm$0.1 & 73.8$\pm$0.1 & 72.0$\pm$0.1 & \textbf{80.0$\pm$0.1} & 73.8$\pm$0.1 \\
			\multirow{2}{*}{Task 12} & ACC & 58.6$\pm$0.1 & 60.5$\pm$0.1 & \textbf{68.0$\pm$0.0} & 55.5$\pm$0.2 & 63.0$\pm$0.0 & 63.5$\pm$0.1 & 64.8$\pm$0.2 & \underline{66.0$\pm$0.3} \\
			& AUC & 66.6$\pm$0.0 & 73.2$\pm$0.0 & \textbf{77.6$\pm$0.1} & 62.4$\pm$0.3 & 74.7$\pm$0.1 & 74.9$\pm$0.2 & 74.5$\pm$0.1 & \underline{75.9$\pm$0.2} \\
			\multirow{2}{*}{Task 13} & ACC & 80.7$\pm$0.1 & 80.4$\pm$0.1 & 82.0$\pm$0.1 & 83.0$\pm$0.2 & 86.4$\pm$0.1 & 87.1$\pm$0.1 & \textbf{90.1$\pm$0.1} & \underline{89.3$\pm$0.1} \\
			& AUC & 90.6$\pm$0.1 & 91.5$\pm$0.1 & 88.9$\pm$0.1 & 90.2$\pm$0.1 & 92.1$\pm$0.1 & \underline{94.5$\pm$0.1} & 94.3$\pm$0.1 & \textbf{95.1$\pm$0.0} \\
			\multirow{2}{*}{Task 14} & ACC & \textbf{69.2$\pm$0.1} & 54.5$\pm$0.2 & 56.6$\pm$0.0 & 58.5$\pm$0.0 & 58.1$\pm$0.0 & 59.6$\pm$0.3 & 58.7$\pm$0.1 & \underline{63.6$\pm$0.1} \\
			& AUC & \textbf{75.2$\pm$0.0} & 67.8$\pm$0.0 & 66.9$\pm$0.0 & 69.0$\pm$0.0 & 65.1$\pm$0.1 & 66.2$\pm$0.2 & 74.1$\pm$0.1 & \underline{74.3$\pm$0.0} \\
			\bottomrule
	\end{tabular}}
	
	\label{result_knn}

\end{table*}

\begin{table*}[]
	\centering
    \caption{Results on linear probing ($k$NN parameter) setting for molecular classification tasks without a validation set. The best score is in bold, and the second-best is underlined.}
	{
		\begin{tabular}{@{}llllllllll@{}}
			\toprule
			Task  & Metric  &\multicolumn{1}{c}{Mean-pool}&\multicolumn{1}{c}{CHIEF} &\multicolumn{1}{c}{PRISM}&\multicolumn{1}{c}{GigaPath}&\multicolumn{1}{c}{TANGLE} &\multicolumn{1}{c}{FEATHER}& \multicolumn{1}{c}{TITAN} & \multicolumn{1}{c}{CARE}\\ \midrule
			\multirow{2}{*}{Task 15} & ACC & \underline{55.2$\pm$0.1} & 54.4$\pm$0.2 & 50.9$\pm$0.1 & 54.7$\pm$0.1 & 53.3$\pm$0.2 & 53.2$\pm$0.1 & \textbf{55.3$\pm$0.1} & 54.3$\pm$0.1 \\
			& AUC & 59.4$\pm$0.3 & 56.1$\pm$0.3 & 51.2$\pm$0.4 & 56.7$\pm$0.3 & 57.5$\pm$0.4 & 59.2$\pm$0.4 & \underline{61.3$\pm$0.2} & \textbf{61.3$\pm$0.3} \\
			\multirow{2}{*}{Task 16} & ACC & 55.9$\pm$0.4 & 64.1$\pm$0.6 & 57.5$\pm$0.3 & 56.2$\pm$0.5 & \textbf{66.8$\pm$0.9} & 54.5$\pm$0.2 & 65.0$\pm$0.7 & \underline{65.5$\pm$0.8} \\
			& AUC & 67.6$\pm$1.3 & \textbf{78.5$\pm$1.0} & 65.0$\pm$0.7 & 67.1$\pm$1.8 & 74.8$\pm$1.4 & 59.5$\pm$1.0 & 74.5$\pm$1.4 & \underline{76.9$\pm$1.0} \\
			

			\multirow{2}{*}{Task 17} & ACC & 55.1$\pm$0.7 & \textbf{61.1$\pm$0.8} & 54.5$\pm$0.6 & 56.0$\pm$0.6 & 56.1$\pm$0.6 & 54.0$\pm$0.4 & \underline{59.4$\pm$0.7} & 59.0$\pm$0.8 \\
			& AUC & 65.5$\pm$1.0 & 67.5$\pm$1.1 & \underline{67.5$\pm$1.4} & 63.8$\pm$1.1 & 63.4$\pm$1.2 & 62.3$\pm$1.9 & 66.9$\pm$1.1 & \textbf{71.4$\pm$1.0} \\
			\multirow{2}{*}{Task 18} & ACC & 65.8$\pm$0.9 & \textbf{68.3$\pm$0.8} & 66.5$\pm$1.0 & \underline{66.8$\pm$0.6} & 64.7$\pm$0.8 & 55.7$\pm$0.3 & 59.8$\pm$0.6 & 66.5$\pm$0.6 \\
			& AUC & 72.6$\pm$1.3 & \textbf{77.3$\pm$1.2} & \underline{76.4$\pm$1.4} & 73.7$\pm$1.6 & 75.9$\pm$1.5 & 59.0$\pm$1.8 & 61.6$\pm$1.3 & 67.9$\pm$1.4 \\
			\multirow{2}{*}{Task 19} & ACC & \textbf{51.9$\pm$0.2} & 50.0$\pm$0.1 & 50.2$\pm$0.1 & 50.1$\pm$0.1 & 50.7$\pm$0.1 & 49.9$\pm$0.0 & \underline{51.4$\pm$0.1} & 50.5$\pm$0.2 \\
			& AUC & 63.7$\pm$3.5 & 50.5$\pm$2.9 & 63.8$\pm$2.0 & 58.5$\pm$3.4 & \underline{65.4$\pm$2.8} & \textbf{68.9$\pm$1.8} & 64.5$\pm$2.6 & 63.2$\pm$2.1 \\
			\multirow{2}{*}{Task 20} & ACC & 47.7$\pm$0.7 & 48.9$\pm$1.1 & 46.0$\pm$0.5 & 44.5$\pm$0.9 & \underline{52.0$\pm$1.0} & \textbf{52.2$\pm$0.8} & 43.0$\pm$0.7 & 47.2$\pm$1.0 \\
			& AUC & 45.5$\pm$1.3 & 48.0$\pm$1.6 & 48.7$\pm$0.8 & 41.9$\pm$1.6 & \underline{49.4$\pm$1.6} & \textbf{51.9$\pm$1.3} & 41.5$\pm$1.3 & 45.5$\pm$1.5 \\
			
			\multirow{2}{*}{Task 21} & ACC & 49.6$\pm$0.1 & \textbf{51.8$\pm$0.2} & 50.2$\pm$0.0 & 49.9$\pm$0.0 & \underline{51.3$\pm$0.2} & 50.3$\pm$0.1 & 49.8$\pm$0.1 & 49.0$\pm$0.1 \\
			& AUC & 70.8$\pm$1.0 & 67.5$\pm$1.3 & \textbf{76.3$\pm$1.3} & 66.4$\pm$1.6 & \underline{72.0$\pm$1.5} & 69.4$\pm$2.3 & 67.8$\pm$1.7 & 65.7$\pm$1.8 \\
			\multirow{2}{*}{Task 22} & ACC & 54.2$\pm$0.3 & 60.2$\pm$0.7 & 54.8$\pm$0.4 & 55.6$\pm$0.6 & \underline{60.6$\pm$0.5} & 51.5$\pm$0.1 & \textbf{63.1$\pm$0.4} & 58.5$\pm$0.6 \\
			& AUC & 79.7$\pm$1.1 & \underline{87.5$\pm$0.7} & 70.5$\pm$1.6 & 83.8$\pm$1.3 & \textbf{91.1$\pm$0.5} & 66.9$\pm$1.3 & 86.7$\pm$0.8 & 81.2$\pm$1.5 \\
			\multirow{2}{*}{Task 23} & ACC & \textbf{53.1$\pm$0.6} & \underline{52.3$\pm$0.4} & 48.8$\pm$0.1 & 51.3$\pm$0.3 & 50.9$\pm$0.2 & 51.6$\pm$0.3 & 51.7$\pm$0.3 & 48.9$\pm$0.1 \\
			& AUC & 56.0$\pm$2.0 & \underline{63.3$\pm$1.4} & 53.0$\pm$1.5 & 60.3$\pm$2.3 & 61.4$\pm$1.8 & 49.7$\pm$1.3 & \textbf{64.9$\pm$1.5} & 59.2$\pm$1.4 \\
			\multirow{2}{*}{Task 24} & ACC & 49.7$\pm$0.2 & 49.1$\pm$0.0 & \textbf{54.3$\pm$0.6} & 50.0$\pm$0.1 & \underline{53.8$\pm$0.6} & 49.8$\pm$0.0 & 51.6$\pm$0.4 & 51.4$\pm$0.3 \\
			& AUC & 65.8$\pm$2.9 & \underline{74.0$\pm$2.9} & 63.8$\pm$3.8 & 71.5$\pm$2.6 & \textbf{79.9$\pm$1.7} & 54.4$\pm$3.3 & 66.9$\pm$2.4 & 72.2$\pm$3.1 \\
			\multirow{2}{*}{Task 25} & ACC & 49.9$\pm$0.0 & 48.7$\pm$0.1 & \textbf{51.1$\pm$0.3} & 49.2$\pm$0.0 & 49.5$\pm$0.0 & \underline{50.0$\pm$0.0} & 49.9$\pm$0.0 & \underline{50.0$\pm$0.0} \\
			& AUC & 53.1$\pm$2.1 & 50.7$\pm$3.8 & \textbf{65.8$\pm$4.2} & 47.2$\pm$3.4 & 50.4$\pm$2.4 & 57.1$\pm$3.1 & \underline{58.2$\pm$2.8} & 55.0$\pm$2.7 \\
			\multirow{2}{*}{Task 26} & ACC & 70.0$\pm$0.6 & 62.8$\pm$0.6 & 63.4$\pm$0.4 & 68.5$\pm$0.7 & 67.2$\pm$0.5 & \textbf{71.5$\pm$0.5} & 67.5$\pm$0.4 & \underline{70.1$\pm$0.7} \\
			& AUC & 75.0$\pm$0.9 & 68.7$\pm$1.0 & 70.8$\pm$0.6 & 74.9$\pm$0.9 & 74.5$\pm$0.9 & \underline{75.1$\pm$0.8} & \textbf{77.7$\pm$0.6} & 73.7$\pm$0.8 \\
			\multirow{2}{*}{Task 27} & ACC & 64.5$\pm$0.7 & 56.1$\pm$0.6 & 57.7$\pm$0.4 & \underline{66.8$\pm$0.8} & 59.3$\pm$0.9 & 65.0$\pm$0.6 & \textbf{68.1$\pm$0.8} & 65.9$\pm$0.7 \\
			& AUC & 71.5$\pm$0.9 & 63.9$\pm$0.9 & 63.4$\pm$0.7 & 74.3$\pm$1.0 & 66.5$\pm$1.3 & 74.2$\pm$0.7 & \textbf{75.9$\pm$0.8} & \underline{75.9$\pm$0.8} \\
			\multirow{2}{*}{Task 28} & ACC & 60.5$\pm$0.6 & 64.1$\pm$0.7 & 64.2$\pm$0.6 & 61.0$\pm$0.7 & \underline{69.3$\pm$0.7} & 62.3$\pm$0.6 & 67.8$\pm$1.2 & \textbf{71.7$\pm$1.4} \\
			& AUC & 82.2$\pm$1.0 & 80.4$\pm$0.8 & 80.0$\pm$1.5 & 82.5$\pm$0.7 & \underline{86.6$\pm$0.6} & 76.8$\pm$1.5 & 84.2$\pm$1.2 & \textbf{87.8$\pm$0.9} \\
			\bottomrule
	\end{tabular}}
	
	\label{result_knn2}

\end{table*}
\begin{itemize}[itemsep=0pt]
\item \textbf{Task 1: EBRAINS-corase.} A coarse-grained 12-class brain tumor subtyping task on the EBRAINS dataset.

\item \textbf{Task 2: EBRAINS-fine.} A fine-grained 30-class brain tumor subtyping benchmark derived from the EBRAINS dataset.

\item \textbf{Task 3: IMP-grading.} On the public IMP-CRS 2024 dataset, each colorectal WSI is weakly labeled as non-neoplastic, low-grade, or high-grade, forming a three-class slide-level tumor grading task.

\item \textbf{Task 4: Local-LUNG-subtype.} Using the Local-LUNG dataset, this task performs three-class WSI-level lung cancer subtype classification. The cohort includes 380 lung adenocarcinoma, 64 small cell lung carcinoma, and 80 lung squamous cell carcinoma cases.

\item \textbf{Task 5: Combine-RCC-subtype.} We construct a three-class RCC subtype classification benchmark by combining cases from the CPTAC-CCRCC cohort and the DHMC-RCC dataset. The resulting cohort contains 713 clear cell RCC, 80 papillary RCC, and 23 chromophobe RCC WSIs.

\item \textbf{Task 6: Combine-LUNG-subtype.} This binary lung cancer subtype task pools WSIs from Local-LUNG, CPTAC-LUAD, CPTAC-LSCC, and DHMC-LUNG into a LUAD-versus-LUSC slide-level benchmark, with 1,660 lung adenocarcinoma (LUAD) and 1,159 lung squamous cell carcinoma (LUSC) WSIs.

\item \textbf{Task 7: Cross-LUNG-fine.} A three-class lung adenocarcinoma growth pattern classification task, with acinar, lepidic, and solid as the predominant histologic patterns. Models are trained and validated on Local-LUNG and evaluated on DHMC-LUNG as an external test set to assess cross-cohort generalization.

\item \textbf{Task 8: MUT-BAP1.} On the MUT-HET-RCC cohort, this binary task predicts BAP1 mutation status from H\&E-stained RCC WSIs, with 162 wild-type and 1,130 mutant slides among 1,292 cases.

\item \textbf{Task 9: MUT-PBRM1.} Also on the MUT-HET-RCC cohort, this binary slide-level task predicts PBRM1 mutation status from H\&E-stained RCC WSIs, including 670 wild-type and 622 mutant slides.

\item \textbf{Task 10: BCNB-ER.} A binary slide-level ER status prediction task on the BCNB cohort, with 219 ER-negative and 809 ER-positive H\&E-stained WSIs.

\item \textbf{Task 11: BCNB-HER2.} On the same cohort, BCNB-HER2 predicts HER2 status at the slide level, comprising 758 HER2-negative and 270 HER2-positive WSIs.

\item \textbf{Task 12: BCNB-PR.} BCNB-PR frames progesterone receptor (PR) status prediction from H\&E-stained WSIs as a binary slide-level task (260 PR-negative and 768 PR-positive slides).

\item \textbf{Task 13: EBRIANS-IDH.} On H\&E-stained glioma WSIs from the EBRAINS dataset, this binary slide-level task predicts isocitrate dehydrogenase (IDH) mutation status (540 IDH–wild-type and 333 IDH–mutant slides).

\item \textbf{Task 14: Cross-MUT-BAP1.} Cross-MUT-BAP1 evaluates cross-cohort generalization for BAP1 mutation prediction by training and validating on MUT-HET-RCC and testing on CPTAC-CCRCC for binary slide-level BAP1 status.

\item \textbf{Task 15: SR386-RAS.} Defined on the SR386 cohort with linked genetic markers, SR386-RAS is a binary slide-level RAS mutation prediction task, including 251 RAS wild-type and 138 RAS mutant cases.

\item \textbf{Task 16: CPTAC-GBM-TP53.} Using WSIs from the CPTAC-GBM cohort, this task performs binary slide-level prediction of TP53 mutation status (162 TP53 wild-type and 80 TP53 mutant cases).

\item \textbf{Task 17: CPTAC-BRCA-HER2.} On H\&E-stained WSIs from the CPTAC-BRCA cohort, CPTAC-BRCA-HER2 formulates binary slide-level HER2 status prediction, with 50 HER2-negative and 39 HER2-positive breast cancer cases.

\item \textbf{Task 18: CPTAC-BRCA-PIK3CA.} CPTAC-BRCA-PIK3CA uses H\&E-stained WSIs from CPTAC-BRCA for binary slide-level PIK3CA mutation prediction (81 PIK3CA wild-type and 34 PIK3CA mutant cases).

\item \textbf{Task 19: CPTAC-CCRCC-BAP1.} On H\&E-stained WSIs from the CPTAC-CCRCC cohort, this task predicts BAP1 mutation status at the slide level, with 202 BAP1 wild-type and 43 BAP1 mutant cases.

\item \textbf{Task 20: CPTAC-CCRCC-PBRM1.} CPTAC-CCRCC-PBRM1 defines binary slide-level PBRM1 mutation prediction on the same cohort, including 123 PBRM1 wild-type and 122 PBRM1–mutant cases.

\item \textbf{Task 21: CPTAC-COAD-APC.} Using H\&E-stained WSIs from the CPTAC-COAD cohort, CPTAC-COAD-APC defines binary slide-level APC mutation prediction (22 APC–wild-type and 76 APC–mutant cases).

\item \textbf{Task 22: CPTAC-COAD-MSI.} CPTAC-COAD-MSI frames microsatellite instability (MSI) status prediction on CPTAC-COAD H\&E WSIs as a binary slide-level task, with 81 MSI-stable and 25 MSI-high cases.

\item \textbf{Task 23: CPTAC-COAD-PIK3CA.} On the same cohort, CPTAC-COAD-PIK3CA performs binary slide-level PIK3CA mutation prediction (73 wild-type and 25 mutant cases).

\item \textbf{Task 24: CPTAC-COAD-SETD1B.} CPTAC-COAD-SETD1B uses H\&E-stained WSIs from CPTAC-COAD for binary slide-level SETD1B mutation status prediction, including 81 SETD1B–wild-type and 17 SETD1B–mutant cases.

\item \textbf{Task 25: CPTAC-HNSCC-CASP8.} For CPTAC-HNSCC-CASP8, H\&E-stained WSIs from the CPTAC-HNSCC cohort are used to predict CASP8 mutation status at the slide level in a binary setting (222 CASP8–wild-type and 33 CASP8–mutant cases).

\item \textbf{Task 26: CPTAC-LUAD-EGFR.} CPTAC-LUAD-EGFR is a binary slide-level EGFR mutation prediction task on H\&E-stained WSIs from the CPTAC-LUAD cohort, with 211 EGFR–wild-type and 113 EGFR–mutant cases.

\item \textbf{Task 27: CPTAC-LUAD-KRAS.} Similarly, this task targets KRAS mutation status on CPTAC-LUAD H\&E WSIs, formulating a binary slide-level prediction task with 200 KRAS wild-type and 112 KRAS mutant cases.

\item \textbf{Task 28: CPTAC-LUAD-STK11.} CPTAC-LUAD-STK11 defines binary slide-level STK11 mutation prediction on the CPTAC-LUAD cohort, comprising 266 STK11–wild-type and 58 STK11–mutant cases.

\item \textbf{Task 29: CPTAC-CCRCC-OS.} CPTAC-CCRCC-OS uses H\&E-stained clear cell RCC WSIs from the CPTAC-CCRCC collection and corresponding overall survival (OS) time and event data from cBioPortal to define a slide-level OS risk prediction task for assessing model prognostic performance.

\item \textbf{Task 30: CPTAC-HNSCC-OS.} On H\&E-stained WSIs from the CPTAC-HNSCC cohort with associated OS time and event annotations, CPTAC-HNSCC-OS formulates a slide-level OS risk prediction benchmark to evaluate both prognostic performance and cross-cohort generalization.

\item \textbf{Task 31: CPTAC-LUAD-OS.} CPTAC-LUAD-OS constructs a slide-level OS risk prediction task on H\&E-stained WSIs from the CPTAC-LUAD cohort, paired with survival time and censoring status from cBioPortal, to assess model prognostic performance and robustness.

\item \textbf{Task 32: CPTAC-PDA-OS.} Using H\&E-stained whole-slide images of pancreatic ductal adenocarcinoma (PDA) from the CPTAC-PDA cohort and OS time with censoring information from the corresponding cBioPortal study, CPTAC-PDA-OS defines a slide-level OS risk prediction task for evaluating prognostic performance and generalization.

\item \textbf{Task 33: SR386-OS.} SR386-OS leverages 427 H\&E-stained WSIs from the SR386 cohort with five-year OS follow-up and event (censoring) status to set up a slide-level OS risk prediction task, used to benchmark model prognostic accuracy and generalization ability.
\end{itemize} 
\begin{table*}[]
	\centering
    \caption{Survival prediction experimental results. All results are obtained under the linear-probing (Linear) setting. The best score is in bold, and the second-best is underlined.}
    {
		\begin{tabular}{@{}lllllllll@{}}
			\toprule
			Task   &\multicolumn{1}{c}{Mean-pool}&\multicolumn{1}{c}{CHIEF} &\multicolumn{1}{c}{PRISM}&\multicolumn{1}{c}{GigaPath}&\multicolumn{1}{c}{TANGLE} &\multicolumn{1}{c}{FEATHER}& \multicolumn{1}{c}{TITAN} & \multicolumn{1}{c}{CARE}\\ \midrule

			{Task 29}  & 48.4$\pm${16.5}&	53.2$\pm${15.4}&	38.0$\pm${20.0}&	55.8$\pm${4.7}&	46.6$\pm${8.3}	&\underline{56.6$\pm$15.3}&	40.5$\pm${18.9}&	\textbf{63.0$\pm$8.9}\\
			Task 30 &53.8$\pm${15.8}&	\underline{58.7$\pm$8.7}&	39.9$\pm${9.9}&	57.3$\pm${11.1}&	48.6$\pm${23.6}	&57.6$\pm${12.4}&	38.9$\pm${12.6}	&\textbf{65.9$\pm$18.0} \\
			{Task 31} & 48.4$\pm${13.8}&	\underline{56.3$\pm$9.6}&	38.0$\pm${21.8}&	\textbf{56.9$\pm$16.1}&	50.8$\pm${9.2}&	46.4$\pm${16.0}&	54.6$\pm${20.6}&	47.5$\pm${11.9} \\
			{Task 32} &42.2$\pm${7.6}&	50.3$\pm${10.2}&	48.3$\pm${8.7}&	46.8$\pm${8.9}&	52.2$\pm${8.1}&	\textbf{61.3$\pm$10.5}&	43.5$\pm${6.6}&	\underline{56.6$\pm$10.1} \\
			{Task 33} & 49.7$\pm${4.8}&	53.0$\pm${7.3}&	46.2$\pm${10.2}&	\underline{62.0$\pm$7.5}&	50.8$\pm${6.0}&	47.4$\pm${6.2}&	\textbf{58.5$\pm$6.5}&	57.1$\pm${3.1} \\
			
			\bottomrule
	\end{tabular}}
	\label{result_os}
\end{table*}
\begin{figure}
  \centering
  \includegraphics[width=\linewidth]{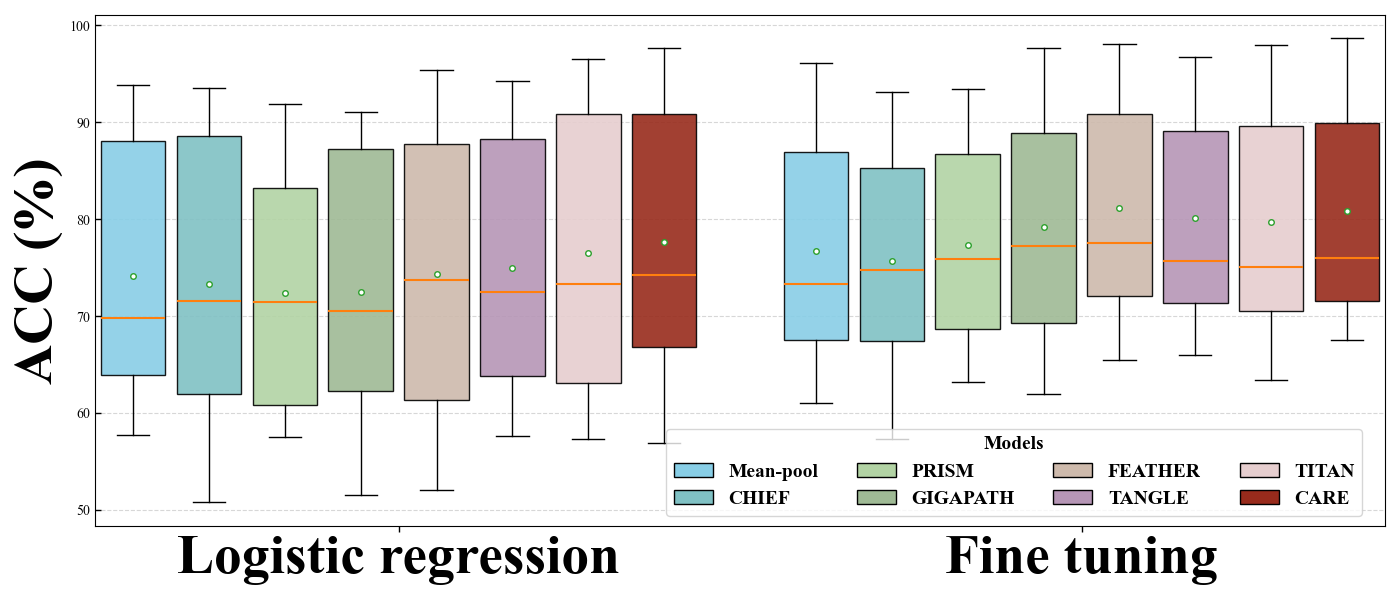}
  \caption{Experimental results of ACC box plots for different models under different experimental settings. Each box plot aggregates results from all morphology-classification and molecular-classification tasks. The horizontal line inside each box indicates the median for that model across tasks. The green circle denotes the mean.}
  \label{acc_box}
\end{figure}

In the above experiments, we group Tasks 1–7 as morphological classification tasks, which are closely related to histologic subtype information in pathology images. Tasks 8–14 are categorized as molecular classification tasks for gene mutation prediction. Tasks 15–28 are molecular classification tasks without a validation set, since the corresponding datasets contain relatively few samples. Tasks 29–33 are categorized as survival analysis tasks for overall survival risk prediction.

\section{Description of Slide-Level Baseline Models}
In this paper, we compare our method with several state-of-the-art slide-level foundation models. Their details are described below.
\begin{itemize}
	\item \textbf{CHIEF \cite{chief}.} CHIEF is an attention-based multiple instance learning slide encoder, which fused with a CLIP text embedding of the WSI’s anatomical site. It is weakly supervised pretrained on 60,530 H\&E WSIs from 19 anatomical sites, using slide-level labels such as anatomical site, cancer type, genomic alterations and prognostic outcomes as multi-task targets.
	\item \textbf{PRISM \cite{prism}.} PRISM is a multimodal vision–language slide-level foundation model that takes H\&E WSIs and pairs them with clinical pathology reports, using a perceiver-based slide encoder in a CoCa-style architecture to jointly encode image and text. It is pretrained on 587,196 WSIs, using contrastive and generative objectives that align slide embeddings with free-text diagnostic reports.
    \item \textbf{GigaPath \cite{xu2024whole}.} GigaPath is a whole-slide vision transformer that treats a WSI as a very long sequence of image tiles. It is a LongNet-based slide encoder with masked-autoencoder pretraining models global context across 171,189 WSIs.
    \item \textbf{FEATHER \cite{feather}.} FEATHER is a lightweight image-only slide-level models that use an attention-based multiple instance learning (ABMIL) framework on top of frozen patch encoders such as  CONCH v1.5 to aggregate patch embeddings into WSI representations. It is supervised-pretrained on the PC-108 pan-cancer morphological classification task about 24,000 H\&E WSIs—using 108-way slide-level morphology labels as targets.
    \item \textbf{TANGLE \cite{tangle}.} TANGLE is a multimodal foundation model that uses a slide encoder together with a gene-expression encoder, and aligns the two modalities via CLIP-style contrastive learning between WSI embeddings and bulk RNA-seq expression embeddings. It is pan-cancer pretrained on slide–expression pairs from all TCGA cohorts, where the main supervision signal comes from paired bulk transcriptomic profiles rather than discrete diagnostic or mutation labels. Notably, in this paper we use the weights of TANGLE v2.
    \item \textbf{TITAN \cite{titan}.} TITAN is a multimodal whole-slide vision–language foundation model that uses a transformer slide encoder, together with a text encoder in a CoCa-style image–text framework to jointly represent H\&E WSIs and pathology reports. It is pretrained on 335,645 WSIs using a three-stage scheme: visual self-supervised learning on slides, followed by vision–language contrastive and captioning objectives that align slide embeddings with 423,122 synthetic captions.
\end{itemize}

\begin{figure*}
  \centering
  \includegraphics[width=\linewidth]{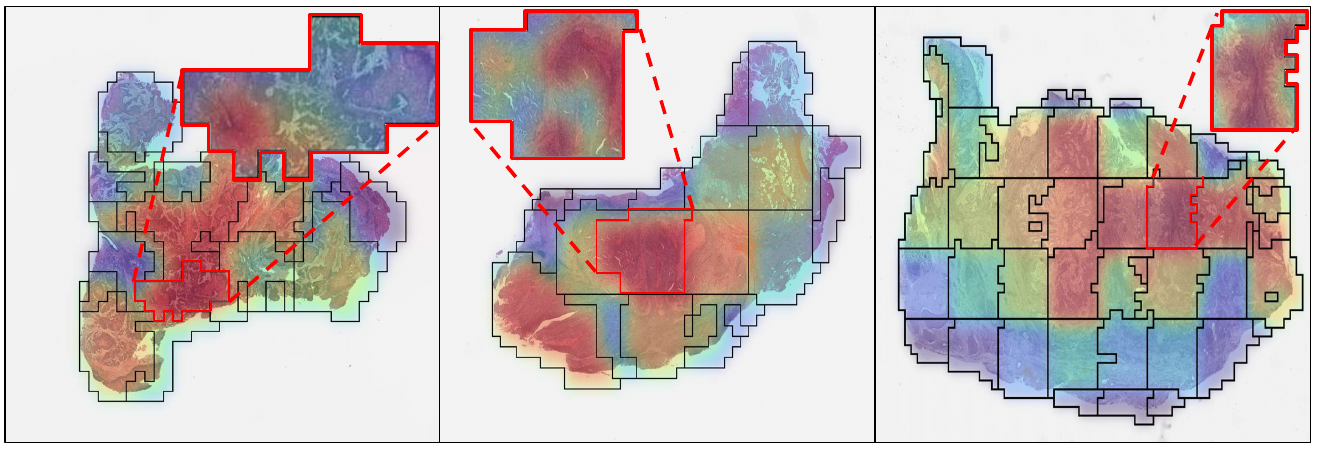}
  \caption{Heatmap visualization of CARE patches within the ROI. The WSIs are sourced from the CPTAC-HNSCC dataset. Warmer colors (red) indicate stronger patch attention.}
  \label{vis_region}
\end{figure*}

Moreover, we provide a detailed comparison between the above models and CARE in terms of their architectures and pretraining data, as shown in \cref{flop}. Here, ``FLOPs'' denotes the vision-inference FLOPs of each model computed using the fvcore toolkit, and ``Para'' denotes the number of trainable parameters involved in a single forward pass from patch features to WSI features, excluding the parameters of the patch encoder. On the one hand, models that use ABMIL or variants of ABMIL as their backbone are lightweight, with low parameter counts and FLOPs, and therefore do not fall into the same regime as foundation models whose defining characteristic is large-scale pretraining. Among the remaining models, CARE has the lowest number of parameters and FLOPs, which accelerates the computation of slide embeddings. On the other hand, CARE is pretrained on only about one tenth of the data used by strong baselines such as TITAN and PRISM, yet still achieves impressive performance, highlighting the advantages of its architecture and pretraining strategy.

\section{Implementation Details for Downstream Task Evaluation}
\noindent\textbf{Implementation Details}. All datasets are preprocessed with the CLAM toolkit \cite{clam} using a unified pipeline. For Tasks 1–14, we evaluate three settings: logistic regression, k-nearest neighbors ($k$NN), and fine tuning. For Tasks 15–28, the datasets are relatively small. We therefore split the data only into training and test sets without a separate validation set for hyperparameter search, and report results for logistic regression and KNN with fixed hyperparameters. For Tasks 29–33 (survival analysis), we again use a train–test split only and apply a single linear layer on top of slide features to map them to logits.
 
In the logistic regression setting, we use the ``lbfgs'' solver. When a validation set is available, we perform a logarithmic grid search over 45 log–uniform values of the regularization parameter (C) in the range $[10^{-6}, 10^{5}]$. When no validation set is available, we fix $C = 0.5$. In the $k$NN setting, if a validation set exists, we search over ($k \in \{3, 5, 7, 9, 11, 13, 15, 17, 19, 23\}$). If no validation set is available, we choose $k = \sqrt{N}$, where $N$ is the number of training samples. In the fine-tuning setting for Tasks 1–14, we attach a linear layer on top of each foundation model to map slide features to logits. The slide-level backbone is initialized from its corresponding pretrained weights, while the linear layer is randomly initialized, and all parameters are fine tuned with a learning rate of $2\times 10^{-5}$. All downstream evaluation tasks are trained and tested on a single NVIDIA RTX 4090 GPU.

\noindent\textbf{Evaluation Metric}. For Tasks 1–28, we report balanced accuracy (ACC) as the primary metric. For Tasks 1–5 and Task 7, which are multi-class classification problems, we additionally report the macro-F1 (F1) score. For Task 6 and Tasks 8–28, which are binary classification problems, we additionally report the AUROC (AUC). We use F1 for multi-class tasks because it directly reflects per-class precision–recall trade-offs under class imbalance, whereas for binary tasks AUROC is preferred as it summarizes the model’s overall threshold-independent discrimination between positive and negative cases. For Tasks 29–33, which are survival analysis tasks, we report the concordance index (C-index). We run 50 Monte Carlo cross-validation splits for Tasks 15–28 and 5 splits for the remaining tasks.


\begin{table*}[]
	\centering
    \caption{Results on fine tuning setting. Since Tasks 15 to 33 do not have a validation set, the fine-tuning experiments were only evaluated on Tasks 1 to 14. The best score is in bold, and the second-best is underlined.}
	{
		\begin{tabular}{@{}llllllllll@{}}
			\toprule
			Task  & Metric  &\multicolumn{1}{c}{Mean-pool}&\multicolumn{1}{c}{CHIEF} &\multicolumn{1}{c}{PRISM}&\multicolumn{1}{c}{GigaPath}&\multicolumn{1}{c}{TANGLE} &\multicolumn{1}{c}{FEATHER}& \multicolumn{1}{c}{TITAN} & \multicolumn{1}{c}{CARE}\\ \midrule
			
			\multirow{2}{*}{Task 1} & ACC & 82.6$\pm$0.0 & 70.6$\pm$0.1 & 78.9$\pm$0.1 & 85.2$\pm$0.0 & 84.5$\pm$0.1 & 88.0$\pm$0.0 & \underline{88.3$\pm$0.1} & \textbf{88.9$\pm$0.1} \\
			& F1 & 83.9$\pm$0.0 & 75.5$\pm$0.1 & 85.0$\pm$0.1 & 90.0$\pm$0.0 & 88.6$\pm$0.0 & 89.4$\pm$0.0 & \textbf{92.0$\pm$0.0} & \underline{91.7$\pm$0.0} \\
			\multirow{2}{*}{Task 2} & ACC & 67.9$\pm$0.1 & 57.4$\pm$0.2 & 63.5$\pm$0.1 & 71.8$\pm$0.0 & 71.3$\pm$0.0 & \underline{74.6$\pm$0.1} & \textbf{75.9$\pm$0.0} & 73.0$\pm$0.1 \\
			& F1 & 68.7$\pm$0.1 & 54.4$\pm$0.2 & 66.2$\pm$0.1 & 73.8$\pm$0.0 & 72.8$\pm$0.0 & 74.9$\pm$0.0 & \textbf{78.3$\pm$0.0} & \underline{75.1$\pm$0.1} \\
			\multirow{2}{*}{Task 3} & ACC & 88.5$\pm$0.0 & 93.1$\pm$0.0 & 93.4$\pm$0.0 & \textbf{93.9$\pm$0.0} & \underline{93.5$\pm$0.0} & 92.8$\pm$0.0 & 93.2$\pm$0.0 & 93.2$\pm$0.0 \\
			& F1 & 87.7$\pm$0.0 & 92.9$\pm$0.0 & 93.4$\pm$0.0 & \textbf{93.9$\pm$0.0} & \underline{93.6$\pm$0.0} & 92.8$\pm$0.0 & 92.5$\pm$0.0 & 92.9$\pm$0.0 \\
			\multirow{2}{*}{Task 4} & ACC & 91.0$\pm$0.2 & 85.6$\pm$0.1 & 92.6$\pm$0.1 & 94.6$\pm$0.0 & 95.0$\pm$0.1 & \underline{97.1$\pm$0.1} & \textbf{97.9$\pm$0.0} & 95.1$\pm$0.1 \\
			& F1 & 94.0$\pm$0.1 & 90.9$\pm$0.0 & 95.6$\pm$0.1 & 96.1$\pm$0.0 & 97.2$\pm$0.0 & \underline{97.4$\pm$0.1} & \textbf{98.2$\pm$0.1} & 97.1$\pm$0.0 \\
			\multirow{2}{*}{Task 5} & ACC & 96.1$\pm$0.1 & 92.7$\pm$0.1 & 89.0$\pm$0.2 & 97.7$\pm$0.1 & 96.8$\pm$0.1 & \underline{98.1$\pm$0.1} & 97.2$\pm$0.1 & \textbf{98.7$\pm$0.0} \\
			& F1 & 96.5$\pm$0.0 & 97.0$\pm$0.0 & 97.2$\pm$0.0 & \underline{99.2$\pm$0.0} & 98.8$\pm$0.0 & \textbf{99.3$\pm$0.0} & 99.0$\pm$0.0 & 98.6$\pm$0.0 \\
			\multirow{2}{*}{Task 6} & ACC & 86.2$\pm$0.0 & 85.5$\pm$0.0 & 86.1$\pm$0.1 & 89.1$\pm$0.0 & \underline{89.1$\pm$0.0} & 87.9$\pm$0.0 & \textbf{89.5$\pm$0.0} & 88.5$\pm$0.1 \\
			& AUC & 94.1$\pm$0.0 & 93.7$\pm$0.0 & 95.0$\pm$0.0 & 96.0$\pm$0.0 & 96.2$\pm$0.0 & 95.3$\pm$0.0 & \textbf{96.7$\pm$0.0} & \underline{96.5$\pm$0.0} \\
			\multirow{2}{*}{Task 7} & ACC & 66.0$\pm$0.0 & 61.4$\pm$0.0 & 66.2$\pm$0.2 & 61.9$\pm$0.2 & \textbf{72.3$\pm$0.1} & 68.2$\pm$0.1 & \underline{71.8$\pm$0.0} & 71.4$\pm$0.1 \\
			& F1 & 50.5$\pm$0.2 & 51.9$\pm$0.4 & 65.2$\pm$1.0 & 57.4$\pm$1.0 & \underline{68.4$\pm$1.0} & 68.0$\pm$0.3 & \textbf{71.3$\pm$0.5} & 65.4$\pm$2.0 \\
			\multirow{2}{*}{Task 8} & ACC & 74.1$\pm$0.0 & \textbf{77.5$\pm$0.3} & 71.9$\pm$0.1 & 74.8$\pm$0.2 & 70.8$\pm$0.2 & \underline{75.3$\pm$0.2} & 67.2$\pm$0.2 & 71.4$\pm$0.2 \\
			& AUC & 83.4$\pm$0.0 & 83.5$\pm$0.1 & 87.1$\pm$0.1 & 86.3$\pm$0.0 & \underline{87.4$\pm$0.0} & 84.0$\pm$0.1 & 87.2$\pm$0.0 & \textbf{89.0$\pm$0.0} \\
			\multirow{2}{*}{Task 9} & ACC & 67.9$\pm$0.2 & 66.9$\pm$0.1 & 71.0$\pm$0.0 & 68.6$\pm$0.1 & \underline{71.8$\pm$0.1} & 71.3$\pm$0.1 & 70.1$\pm$0.1 & \textbf{72.2$\pm$0.1} \\
			& AUC & 73.5$\pm$0.2 & 74.1$\pm$0.1 & 77.8$\pm$0.0 & 77.3$\pm$0.1 & \textbf{80.9$\pm$0.2} & 78.4$\pm$0.1 & 78.1$\pm$0.1 & \underline{79.3$\pm$0.1} \\
			\multirow{2}{*}{Task 10} & ACC & 72.6$\pm$0.2 & 76.4$\pm$0.0 & 78.1$\pm$0.1 & \underline{79.7$\pm$0.6} & 78.9$\pm$0.2 & \textbf{79.7$\pm$0.1} & 73.5$\pm$0.1 & 76.5$\pm$0.3 \\
			& AUC & 81.5$\pm$0.2 & 83.4$\pm$0.2 & \underline{89.4$\pm$0.1} & \textbf{90.0$\pm$0.2} & 88.6$\pm$0.1 & 87.1$\pm$0.1 & 86.8$\pm$0.0 & 88.0$\pm$0.1 \\
			\multirow{2}{*}{Task 11} & ACC & 61.0$\pm$0.3 & 67.2$\pm$0.1 & 67.9$\pm$0.2 & 68.0$\pm$0.1 & \textbf{70.7$\pm$0.1} & \underline{70.4$\pm$0.1} & 63.5$\pm$0.1 & 67.5$\pm$0.2 \\
			& AUC & 69.4$\pm$0.1 & 73.2$\pm$0.1 & 77.0$\pm$0.1 & \underline{77.6$\pm$0.1} & \textbf{79.4$\pm$0.1} & 77.3$\pm$0.0 & 71.7$\pm$0.3 & 77.6$\pm$0.1 \\
			\multirow{2}{*}{Task 12} & ACC & 67.4$\pm$0.1 & 73.0$\pm$0.1 & 73.6$\pm$0.3 & 71.2$\pm$0.6 & 72.4$\pm$0.3 & \underline{74.8$\pm$0.1} & 74.1$\pm$0.1 & \textbf{75.4$\pm$0.1} \\
			& AUC & 75.4$\pm$0.1 & 78.3$\pm$0.1 & \underline{82.2$\pm$0.0} & 81.5$\pm$0.1 & 82.0$\pm$0.1 & 81.3$\pm$0.1 & 82.1$\pm$0.1 & \textbf{82.9$\pm$0.1} \\
			\multirow{2}{*}{Task 13} & ACC & 87.1$\pm$0.1 & 85.0$\pm$0.1 & 86.9$\pm$0.1 & 88.3$\pm$0.0 & 89.1$\pm$0.1 & \textbf{91.9$\pm$0.1} & 89.7$\pm$0.0 & \underline{90.2$\pm$0.1} \\
			& AUC & 93.9$\pm$0.0 & 92.7$\pm$0.1 & 95.0$\pm$0.0 & 96.0$\pm$0.0 & \underline{96.3$\pm$0.0} & 95.8$\pm$0.1 & 96.2$\pm$0.0 & \textbf{96.5$\pm$0.0} \\
			\multirow{2}{*}{Task 14} & ACC & 66.0$\pm$0.0 & \underline{67.9$\pm$0.6} & 63.2$\pm$0.2 & 63.8$\pm$0.7 & 65.9$\pm$0.2 & 65.5$\pm$0.1 & 64.1$\pm$0.2 & \textbf{69.7$\pm$0.1} \\
			& AUC & 71.5$\pm$0.0 & 77.2$\pm$0.1 & 78.0$\pm$0.1 & \underline{84.5$\pm$0.0} & 83.5$\pm$0.0 & 71.5$\pm$0.1 & 81.3$\pm$0.1 & \textbf{85.0$\pm$0.1} \\
			\bottomrule
	\end{tabular}}
	
	\label{result_fine}
\end{table*}

\section{Experimental Results in Detail}
We present detailed experimental results for 75 experiments on 33 tasks across 9 datasets in \cref{result_logistic,result_logistic_cptac,result_knn,result_knn2,result_os,result_fine}. In the era of foundation models, no single model achieves state-of-the-art performance across all tasks, and each model tends to exhibit unique strengths on specific benchmarks. Across 145 metrics from 75 experiments spanning 33 tasks, CARE attains 39 best and 50 second-best results among all compared methods. Its advantage is particularly pronounced on molecular prediction and survival analysis tasks shown in \cref{sum_result1,sum_result_auc}, likely due to the guidance provided by molecular features during pretraining. Intuitively, performance on molecular prediction tasks remains weaker than on morphologic classification across all models. While existing approaches already meet practical requirements for morphologic classification in clinical deployment, CARE represents a modest yet meaningful step toward deployable models for molecular prediction.

For the survival prediction tasks, we observe that all models exhibit large standard deviations and relatively low performance. On the one hand, survival analysis is intrinsically challenging when relying solely on histopathology images. On the other hand, we only report linear probing results for all foundation models rather than fully fine-tuned performance. Notably, compared with the second-best model, CARE achieves a 7.2\% improvement on Task 30, highlighting its substantial potential for survival analysis. A distinctive advantage of CARE over other models is its ability to explicitly identify regions of interest (ROIs) in pathology slides that are highly associated with survival outcomes.

\begin{table*}
  \caption{Average AUC (or F1) results across task categories. The best score is in bold, and the second-best is underlined.}
  \label{sum_result_auc}
  \centering
  \begin{tabular}{@{}llllllllll@{}}
    \toprule
    Task  & Head  &\multicolumn{1}{c}{Mean-pool}&\multicolumn{1}{c}{CHIEF} &\multicolumn{1}{c}{PRISM}&\multicolumn{1}{c}{GigaPath}&\multicolumn{1}{c}{TANGLE} &\multicolumn{1}{c}{FEATHER}& \multicolumn{1}{c}{TITAN} & \multicolumn{1}{c}{CARE}\\ \midrule

			\multirow{3}{*}{\shortstack{Morph.\\Class.}} &LR& 87.2$\pm$0.0 & 86.6$\pm$0.0 & 84.5$\pm$0.0 & 85.3$\pm$0.0 & 87.1$\pm$0.0 & 87.4$\pm$0.1 & \underline{90.1$\pm$0.0} & \textbf{90.1$\pm$0.0} \\
			& $$k$$NN& 82.2$\pm$0.0 & 80.3$\pm$0.1 & 79.8$\pm$0.0 & 78.0$\pm$0.0 & 83.6$\pm$0.0 & 81.8$\pm$0.0 & \textbf{88.1$\pm$0.0} & \underline{87.3$\pm$0.0} \\
			&FT& 82.2$\pm$0.1 & 79.5$\pm$0.1 & 85.4$\pm$0.2 & 86.6$\pm$0.2 & 87.9$\pm$0.1 & 88.1$\pm$0.1 & \textbf{89.7$\pm$0.1} & \underline{88.2$\pm$0.3} \\
			\midrule
			\multirow{3}{*}{\shortstack{Molecular\\Class.}}&LR& 79.8$\pm$0.1 & 82.5$\pm$0.1 & 82.7$\pm$0.1 & 83.1$\pm$0.1 & \underline{83.5$\pm$0.1} & 82.0$\pm$0.1 & 81.1$\pm$0.2 & \textbf{83.6$\pm$0.1} \\
			&$$k$$NN & 72.5$\pm$0.1 & 73.8$\pm$0.1 & 75.7$\pm$0.1 & 71.3$\pm$0.2 & 75.5$\pm$0.1 & 74.5$\pm$0.1 & \underline{76.7$\pm$0.2} & \textbf{79.1$\pm$0.2} \\
			&FT& 78.3$\pm$0.1 & 80.3$\pm$0.1 & 83.8$\pm$0.1 & 84.7$\pm$0.1 & \underline{85.4$\pm$0.1} & 82.2$\pm$0.1 & 83.3$\pm$0.1 & \textbf{85.5$\pm$0.1} \\
			\midrule
			\multirow{2}{*}{\shortstack{{Molecular}\\$\text{Class.}_V$}}&LR&69.9$\pm$1.4&67.6$\pm$1.6&66.0$\pm$1.7&69.9$\pm$1.4&\underline{71.5$\pm$1.2} &65.3$\pm$1.8& 71.5$\pm$1.2 &\textbf{73.1$\pm$1.3}\\
			&$$k$$NN& 65.9$\pm$1.5 & 66.0$\pm$1.6 & 64.6$\pm$1.5 & 65.6$\pm$1.6 & \textbf{69.2$\pm$1.4} & 63.7$\pm$1.6 & \underline{68.5$\pm$1.4} & 68.2$\pm$1.4 \\
    \bottomrule
  \end{tabular}
\end{table*}

The box-plot results for Tasks 1–14 under different experimental settings are shown in \cref{auc_box,acc_box}. Based on the balanced accuracy, we observe that fine-tuning consistently outperforms linear probing with logistic regression. Considering the mean, median, and interquartile ranges of the box plots, CARE delivers highly competitive performance, demonstrating that its adaptive region modeling remains effective even when pretrained on only one-tenth of the data used by mainstream models.
\section{Interpretability Analysis}

In \cref{vis}, we show a heatmap visualization of the adaptive regions to highlight which regions are more important. Each region is rendered with a uniform color (for better visual appearance, we apply Gaussian blur to smooth the region boundaries). The pathologist provided detailed annotations of the ROIs on the WSI. In the annotated regions, prominent cytologic atypia is present, with deeply staining cytoplasm and numerous mitotic figures. Intercellular bridges are scant, and focal central necrosis is observed within some tumor nests. The ROI identified by CARE (the most intensely highlighted red region) closely matches the pathologist-annotated areas. Moreover, from CARE-iBOT to CARE-RNA and finally to CARE, the ROI attention progressively converges toward these expert-annotated regions.

To further investigate the contribution of different patches within the ROI, we additionally visualize an attention-score heatmap over the patches inside the ROI, shown in \cref{vis_region}. Each subfigure shows the adaptive region scores of the SPF module under the CARE pretrained weights. The zoomed-in inset in each subfigure displays the attention-score heatmap of patches within the ROI, indicating which patches contribute most strongly to the ROI representation.

\end{document}